\documentclass[10pt,journal,cspaper,compsoc]{IEEEtran}
\pdfoutput=1
\usepackage{multirow}
\usepackage{balance}
\usepackage{amsfonts}
\usepackage{amssymb}
\usepackage{graphicx}
\usepackage{subfigure}
\usepackage{epstopdf}
\usepackage[noend,ruled]{algorithm2e}
\usepackage{microtype}

\newtheorem{thm:def}{Definition}
\newtheorem{thm:eg}{Example}
\newtheorem{thm:lem}{Lemma}

\newcommand{\method}{\textsc{DPPred}\xspace}
\newcommand{\ddpmine}{\textsc{DDPMine}\xspace}

\newcommand{\pc}{patient cluster}
\newcommand{\winner}{the Top Solution}
\newcommand{\Winner}{The Top Solution}
\newcommand{\prevchallenge}{the 2012 challenge}
\newcommand{\Prevchallenge}{The 2012 challenge}
\newcommand{\rate}{$\Delta$ALSFRS/$\Delta T$}
\newcommand{\mquote}[1]{``\emph{#1}''}

\begin{document}
\title{
	DPPred: An Effective Prediction Framework with Concise Discriminative Patterns
}
\author{
	Jingbo~Shang,~Meng~Jiang,~Wenzhu~Tong,~Jinfeng~Xiao,~Jian~Peng,~Jiawei~Han,\\the Pooled Resource Open-Access ALS Clinical Trials Consortium$^*$\thanks{\scriptsize $*$ Data used in the preparation of this article were obtained from the Pooled Resource Open-Access ALS Clinical Trials (PRO-ACT) Database. As such, the following organizations and individuals within the PRO-ACT Consortium contributed to the design and implementation of the PRO-ACT Database and/or provided data, but did not participate in the analysis of the data or the writing of this report: (1) Neurological Clinical Research Institute, MGH; (2) Northeast ALS Consortium; (2) Novartis; (3) Prize4Life; (4) Regeneron Pharmaceuticals, Inc.; (5) Sanofi; (6) Teva Pharmaceutical Industries, Ltd.}
\IEEEcompsocitemizethanks{
\IEEEcompsocthanksitem
J. Shang, M. Jiang, W. Tong, J. Xiao, J. Peng, J. Han
are with the Department of Computer Science in University of Illinois at Urbana-Champaign, IL, USA.
E-mail: \{shang7, mjiang89, wtong8, jxiao13, jianpeng, hanj\}@illinois.edu
}
\thanks{}}
\markboth{IEEE Transactions on Knowledge and Data Engineering, Manuscript ID}%
{}
\IEEEcompsoctitleabstractindextext{
\begin{abstract}
In the literature, two series of models have been proposed to address prediction problems including classification and regression.
Simple models, such as generalized linear models, have ordinary performance but strong interpretability on a set of simple features. The other series, including tree-based models, organize numerical, categorical and high dimensional features into a comprehensive structure with rich interpretable information in the data.
In this paper, we propose a novel Discriminative Pattern-based Prediction framework (\method) to accomplish the prediction tasks by taking their advantages of both effectiveness and interpretability.  Specifically, \method adopts the concise discriminative patterns that are on the prefix paths from the root to leaf nodes in the tree-based models. 
\method selects a limited number of the useful discriminative patterns by searching for the most effective pattern combination to fit generalized linear models. 
Extensive experiments show that in many scenarios, \method provides competitive accuracy with the state-of-the-art as well as the valuable interpretability for developers and experts. In particular, taking a clinical application dataset as a case study, our \method outperforms the baselines by using only 40 concise discriminative patterns out of a potentially exponentially large set of patterns.
\end{abstract}

}
\maketitle
\section{Introduction}

Accuracy and interpretability are two desired goals in predictive modeling, including both \emph{classification} and \emph{regression}.
Previous work can be characterized into two lines.  
One line has ordinary performance with strong interpretability on a set of simple features, but meets a serious bottleneck when modeling complex high-order interactions between features, such as linear regression, logistic regression~\cite{hosmer2004applied}, and support vector machine~\cite{suykens1999least}.
The other line consists of models that are more often studied for their high accuracy, for example, tree-based models including random forest~\cite{chen2004using} and gradient boosted trees~\cite{ganjisaffar2011bagging} as well as the neural network models~\cite{krizhevsky2012imagenet}, which model nonlinear relationships with high-order combinations of different features. However, their lower interpretability and high complexity prevent practitioners from deploying in practice~\cite{hosmer2004applied}.
In the real-world scientific and medical applications which require both intuitive understanding of the features and high accuracies, the practitioners are not satisfied with neither line of models, and thus, it is important and challenging to develop an effective prediction framework with high interpretability when dealing with high-order interactions with features.

Many pattern-based models have been proposed in the last decade to construct high-order patterns from the large set of features, including association rule-based methods on categorical data~\cite{ma1998integrating,li2001cmar,yin2003cpar,cong2005mining,wang2005harmony,veloso2006lazy} and frequent pattern-based algorithms on text data~\cite{lodhi2002text,leslie2002spectrum} and graph data~\cite{kudo2004application,deshpande2005frequent}. Recently, a novel series of models, the discriminative pattern-based models~\cite{cheng2007discriminative,cheng2008direct}, have demonstrated their advantages over the traditional models. They prune non-discriminative patterns from the whole set of frequent patterns, however, the number of discriminative patterns used in their classification or regression models is still huge (at the magnitude of thousands). How to select concise discriminative patterns for better interpretability is still an open issue.

To address the above challenges, in this paper, we propose a novel discriminative patterns-based learning framework (\method) that extracts a concise set of discriminative patterns from high-order interactions among features for accurate classification and regression. In \method, first we train tree-based models to generate a large set of high-order patterns. Second, we explore all prefix paths from root nodes to leaf nodes in the tree-based models as our discriminative patterns. Third, we compress the number of discriminative patterns by selecting the most effective pattern combinations that fit into a generalized linear model with high classification accuracy or small regression error. This component of fast and effective pattern extraction enables the strong predictability and interpretability of \method.

Intuitively speaking, \method selects the robust discriminative patterns in multi-tree based models by fitting them into a generalized linear model. Our extensive experiments demonstrate that \method achieves comparable or even better performance when competing with the traditional tree-based models. Besides the effectiveness, we want to highlight that our \method framework is applicable in the real-world tasks where the model storage and computational cost are highly restricted.

\begin{figure}[t]
  \centering
  \includegraphics[width=0.5 \textwidth]{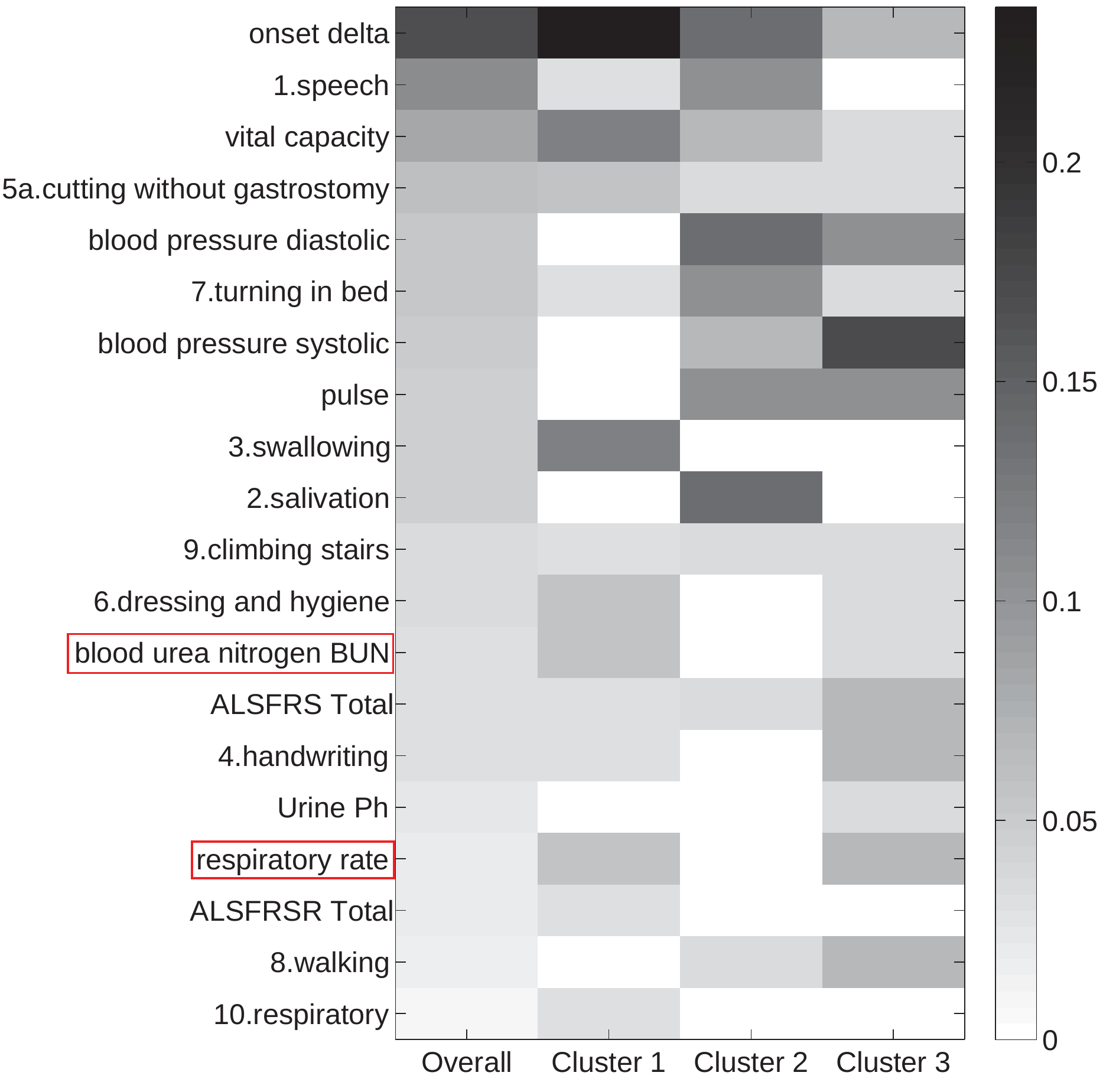}
  \caption{\textit{Two new important factors with the ALS disease that we found with \method.} Among the set of important clinical variables (rows) that \method discovered from the dataset of the Prize4Life Challenge 2012, two highlighted ones have later been experimentally verified that they have extremely high correlations with the ALS disease \cite{zhu2002minocycline,gordon2004placebo,kuffner2015crowdsourced}. The columns are patient clusters.}
  \label{fig:feature_importance}
  \vspace{-0.2in}
\end{figure}

\textit{Discovering robust patterns in the Prize4Life Challenge.}
We apply \method to analyze the prognosis and perform stratification for Amyotrophic Lateral Sclerosis (ALS) patients on the public dataset from the DREAM-Phil Bowen ALS Prediction Prize4Life Challenge 2012. Our \method makes the following achievements.
\begin{itemize}
	\item \method achieves a smaller error, a RMSE of 0.5306, than the method ranked at \#7 with a RMSE of 0.5664. The RMSE of \method is less than 4\% higher than the winner with a RMSE of 0.5113.
	\item The robust discriminative patterns found by our \method are well interpretable, while the other methods including the winner cannot interpret their performances. Note that our \method selects only 40 concise discriminative patterns involving 28 clinical variables from an exponentially large set, while other models used as many as 2 to 3 times variables.
	\item As show in Figure~\ref{fig:feature_importance}, \method discovers two new important clinical factors, the Blood Urea Nitrogen (BUN) and the respiratory rate. These two factors were not found by the top teams in the Challenge but there is indirect experimental and logical evidence for their being actually worth further study~\cite{zhu2002minocycline,gordon2004placebo,kuffner2015crowdsourced}. Also, from the figure we can observe that each patient cluster generates different diagnosis patterns.
\end{itemize}
Our \method accurately predicts the ALS prognosis and systematically identifies clinically-relevant features for the ALS patient stratification in an interpretable manner. The distinct diagnosis patterns can significantly benefit the treatment of the ALS and precision medicine.

It is worthwhile to highlight the advantages of our proposed machine learning framework \method.
\begin{itemize}
    \item \textit{Interpretability.} \method learns a small number of robust discriminative patterns involving high-order interactions among original features.
    \item \textit{Efficiency.} \method compresses multi tree-based models into a low-dimensional generalized linear model, making the online prediction extremely fast.
    \item \textit{Effectiveness.} Experimental results on several real-world datasets demonstrate that \method has comparable or even better performances than the state-of-the-art models on the standard tasks of classification and regression.
    \item \textit{Clinical pattern discovery.} \method has been successfully applied to discover patient clusters and crucial clinical signals for the Amyotrophic Lateral Sclerosis (ALS) disease.
\end{itemize}

The remaining of this paper is organized as follows.
In Section 2 we survey the related work.
In Section 3 we provide the problem definition and our preliminary study.
Section 4 presents our proposed \method framework and the details of its algorithms.
Section 5 reports empirical results on synthetic and real-world datasets.
Section 6 shows our discovery in the prognosis analytic for ALS patients.
Section 7 concludes the study.

\section{Related Work}

In this section we review existing methods that are related to \method, including pattern-based classification models, tree-based models and pattern selection approaches.

\vspace{-0.1in}
\subsection{Pattern-based Classification}

The philosophy of frequent pattern mining has been widely adopted to study the problem of pattern-based classification. Li et al. proposed a classification method CMAR based on multiple class-association rules~\cite{li2001cmar}. Yin et al. extended it to CPAR based on predictive association rules~\cite{yin2003cpar}. Besides the association rules, direct discriminative pattern mining was proposed to generate effective performance~\cite{cheng2007discriminative,cheng2008direct,fan2008direct}. However, these approaches have several serious issues. First, the huge number of frequent patterns leads to expensive computational cost of pattern generation and selection. Second, the number of the selected patterns can be still as large as thousands, which limits the interpretability and causes the inefficiency of the classification model. Third, these models are not capable to address the regression tasks. Moreover, the discretization of continuous variables depends too much on parameter tuning to generate robust performances. Recently, Dong et al.\ proposed to utilize patterns in a different angle, where data are partitioned based on patterns, and complex models are trained independently in different partitions~\cite{sdm16_dong_pattern}. Although this type of pattern aided models sheds lights on a different usage of patterns, the model still lacks of interpretablity.

\vspace{-0.1in}
\subsection{Tree-based Models}
Tree-based models are popular in the classification tasks. Both decision tree and boosted tree models are explainable but quite sensitive to the training data. Traditional ensemble methods using multiple trees, such as random forest~\cite{chen2004using} and gradient boosting decision trees~\cite{friedman2001greedy}, alleviate the over-fitting issue. Ren et al.\ showed that the global refinement could provide better performance because the growth and pruning processes in different trees are independent~\cite{ren2015global}. However, the increased model size of those multi-tree based models sacrifices the interpretability. Our proposed \method is different from this category of models.

There are post-pruning techniques for multi-tree based models to induce new feature spaces. Typically, they encoded each tree as a flat index list and each instance as a binary vector indexed by the trees~\cite{ren2015global,kobetski2011discriminative,ebina2011drop,moosmann2007fast,lou2012intelligible}. Vens et al. transferred the binary vectors into an inner product kernel space using a support vector machine and showed the increase of classification accuracy~\cite{vens2011random}. Furthermore, pairwise interactions have also been studied to fit a two-layer-tree model for accurate classification and regression~\cite{lou2013accurate}.
Though the number of features is reduced by pruning, the dimension of the newly-created feature space is still high due to a large number of constructed trees. For example, in~\cite{ren2015global}, after many efforts on pruning, the model size of the pruned random forest was still at megabytes and thus the prediction was too slow to support real-time applications. Our experimental results will later show that \method delivers comparable results using as few as the top 20 discriminative patterns, which is substantially reduced even compared to the state-of-the-art models.

\vspace{-0.1in}
\subsection{Pattern Selection}
Simply selecting patterns with the highest independent heuristics such as information gain and gini index is limited to very simple tasks due to the redundancy and over-fitting problems~\cite{lee2006information}. Given the labels, i.e., the types for classification or the real numbers for regression, LASSO~\cite{tibshirani1996regression} is widely used in feature selection tasks as well as forward selection~\cite{derksen1992backward}. Due to the relatively large number of candidate discriminative patterns, backward selection is not suitable in our problem setting. Our proposed \method framework adopts the LASSO and forward selection methods to select discriminative patterns. Their performances have been compared and discussed in the experimental section.

\section{Preliminaries}

This section defines the problem as well as the important concepts used throughout this paper.

    \vspace{-0.1in}
    \subsection{Problem Formulation}
        For a prediction task (classification or regression), the data is a set of $n$ examples in a $d$-dimensional feature space together with their labels $(\mathbf{x}_1, y_1), (\mathbf{x}_2, y_2), \ldots, (\mathbf{x}_n, y_n)$, for $\forall i$ ($1$$\le$$i$$\le$$n$), $\mathbf{x}_i \in \mathbb{R}^d$. It is worth noting that the values in the example $\mathbf{x}_i$ can be either continuous (numerical) or discrete (categorical). As categorical features can be transformed into several binary dummy indicators, we can assume $\mathbf{x}_i \in \mathbb{R}^d$ without the loss of generality. The label $y_i$ is either a class (type) indicator or a real number depending on the specific task. In previous pattern-based models, e.g., \ddpmine~\cite{cheng2008direct}, patterns are extracted from categorical values and thus they are only able to handle the continuous variables after careful manual discretization, which is tricky and often requires prior knowledge about the data.

        The goal of our proposed framework \method is to learn a concise model that consists of a small set of discriminative patterns from the training data, which learns and predicts the examples as accurately as possible, i.e., predict the correct class indicator in \textit{classification} tasks and predict close to the true number in \textit{regression} tasks. Formally, given a dataset $\cal{D}$, \method returns a set of $k$ discriminative patterns $\cal{P}$ using a generalized linear model $f(\cdot)$ that minimizes $\sum_{i=1}^n l(f(M(\mathbf{x}_i)), y_i)$, where $l(\cdot, \cdot)$ is the general loss function, $M(\cdot)$ is a mapping function that maps the original feature vector $\mathbf{x}$ to the pattern space using patterns $\cal{P}$.

        \method generates a pool of discriminative patterns within a reasonable size, and selects top-$k$ patterns based on their learning performance on training data, using a generalized linear learning model. Since the number of selected patterns is very limited, these patterns are able to provide informative interpretability with reasonable predictive power. In addition, for the coming testing data, by evaluating only a very small set of the selected discriminative patterns, \method is enabled to make predictions with a generalized linear model efficiently.

        \begin{figure*}[t]
          \centering
          \includegraphics[width = \textwidth]{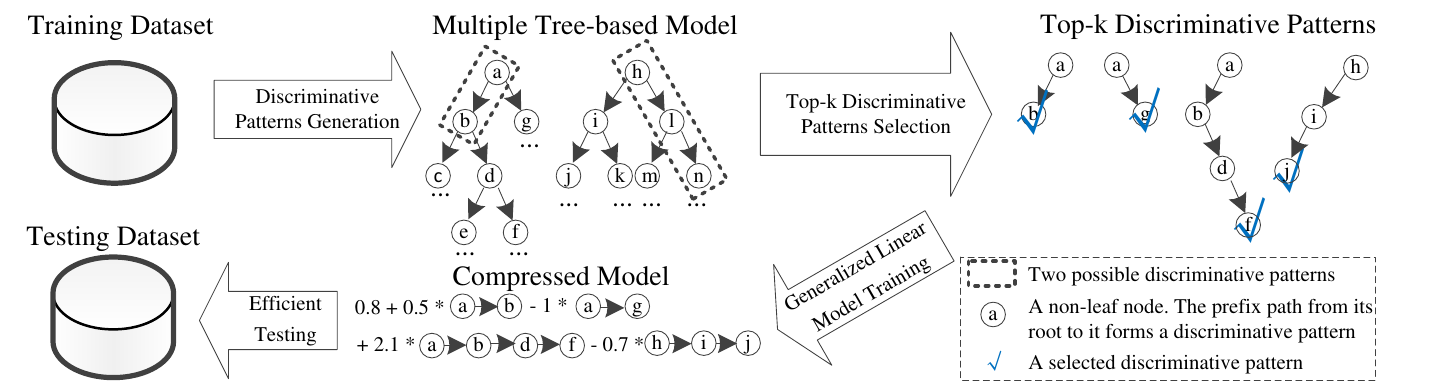}
          \caption{\textit{The overview of our \method framework.} With the training data, the multi-tree based model (e.g., random forest) is trained for discriminative pattern generation. For each tree, all prefix paths from its root to non-leaf nodes are treated as discriminative patterns. After a large pool of discriminative patterns is generated, \method conducts top-$k$ pattern selection to identify the most informative and interpretable patterns. Here $k$ is typically small (20 or 30). Finally, it trains a generalized linear model based on the $2^k$ pattern space representation.}
          \label{fig:framework}
          \vspace{-0.2in}
        \end{figure*}

    \vspace{-0.1in}
    \subsection{Definition}
        First, we define a series of concepts to derive the discriminative patterns. Traditional frequent pattern mining works on categorical data and itemset data, in which discretization is required to deal with continuous variables. Instead of roughly discretizing the numerical values, we adopt the thresholding boolean function in \method.

        \begin{thm:def} \textbf{Condition} is a thresholding boolean function on a specific feature dimension. The condition is in the form of $(x_{\cdot,j} < v)$ or $(x_{\cdot,j} \ge v)$, where $j$ indicates the specific dimension and $v$ is the threshold value. The relational operator in a condition is either $<$ or $\ge$. For any dimension $j$ in features corresponding to binary indicators, we restrict $v$ to be $0.5$.
        \end{thm:def}

        Note that the threshold values in \method are not specified by users beforehand. In previous pattern-based models, e.g., \ddpmine~\cite{cheng2008direct}, the practitioners have to discretize values of continuous variables prior to pattern mining. \method automatically determines these values in the tree model, completely based on the training data without any human interventions.

        \begin{thm:eg} Suppose $\mathbf{x}_i \in \mathbb{R}^{10}$, one possible condition is that $\mathbf{x}_{\cdot, 1} < 0.5$. Another example could be $\mathbf{x}_{\cdot, 2} \ge 0.8$.
        \end{thm:eg}

        We define a pattern as a set of conditions. Formally, we use conjunctions to concatenate different conditions: it is consistent with the prefix path in the decision tree that represents the conjunction of the conditions in the nodes along the path.

        \begin{thm:def} \textbf{Pattern} is a conjunction clause of conditions on specific feature dimensions. Formally, it is defined as follows.
        \[
            (x_{\cdot,j_1} < v_1) \wedge (x_{\cdot,j_2} \ge v_2) \wedge \ldots \wedge (x_{\cdot,j_m} \ge v_m),
        \]
        where $m$ is the number of conditions within this pattern. Different patterns are allowed to have different $m$ values.
        \end{thm:def}

        \begin{thm:eg}
        Suppose $\mathbf{x}_i \in \mathbb{R}^{10}$, one possible pattern is that $(\mathbf{x}_{\cdot, 1} < 18) \wedge (\mathbf{x}_{\cdot, 3} \ge 100) \wedge (\mathbf{x}_{\cdot, 9} < 0.5)$.
        \end{thm:eg}

        Now we define discriminative patterns as follows.
        \begin{thm:def} \textbf{Discriminative Patterns} refer to those patterns which have strong signals on the learning tasks, given the labels of data. For example, a pattern with very high information gain on the classification training data, or a pattern with very small mean square error on the regression training data, is a discriminative pattern.
        \end{thm:def}

        \begin{thm:eg}
        Suppose $\mathbf{x}_i \in \mathbb{R}^{10}$ and the labels are generated as follows.
        \[
            y_i = [(\mathbf{x}_{i, 1} \ge 1) \wedge (\mathbf{x}_{i, 2} < 0)] \vee [(\mathbf{x}_{i, 1} < 18) \wedge (\mathbf{x}_{i, 3} \ge 100)].
        \]
        Both patterns $(\mathbf{x}_{i, 1} \ge 1) \wedge (\mathbf{x}_{i, 2} < 0)$ and $(\mathbf{x}_{i, 1} < 18) \wedge (\mathbf{x}_{i, 3} \ge 100)$ are two of the most discriminative patterns. Similar patterns that contain or have overlaps with these two patterns are also discriminative patterns.
        \end{thm:eg}

        Discriminative patterns have overlapped predictive effects. Specifically, a few discriminative patterns are special cases of other patterns. For example, in the previous example, both patterns $(\mathbf{x}_{i, 1} \ge 1) \wedge (\mathbf{x}_{i, 2} < 0)$ and $(\mathbf{x}_{i, 1} \ge 1) \wedge (\mathbf{x}_{i, 2} < 0) \wedge (\mathbf{x}_{i, 3} < 0)$ indicate a positive label. However, the second pattern only encodes a subset of data points that the first pattern encodes, and thus, it does not provide extra information for the learning process. This common phenomenon shows that roughly taking the top discriminative patterns based on independent heuristics wastes the budget of the number of patterns, when the linear combination of these patterns are not synergistic. Therefore, our \method selects the top-$k$ patterns by their predictive performance to make the selected patterns complementary and compact.

        \begin{thm:def} \textbf{Top-$k$ Patterns} are formalized as a size-$k$ subset of discriminative patterns, which has the best performance (i.e., the highest accuracy in classification tasks or the least rooted mean square error in regression tasks) based on the training data.
        \end{thm:def}

        Here we assume that the training and testing data share the same distribution, which is widely acknowledged in the classification and regression problems. In this case, the accuracy on the testing data is approaching the accuracy on the training data and our model is able to alleviate the over-fitting issue.

        \begin{thm:eg}
        In the last example, the top-$2$ patterns are $\{ (\mathbf{x}_{i, 1} \ge 1) \wedge (\mathbf{x}_{i, 2} < 0), (\mathbf{x}_{i, 1} < 18) \wedge (\mathbf{x}_{i, 3} \ge 100) \}$.
        \end{thm:eg}

\section{Our DPPred Framework}
\label{sec:method}

    This section first presents the overview of \method and then introduces the details of every component in this framework as well as the theoretical time complexity.

    \vspace{-0.1in}
    \subsection{The Overview of \method}
        Figure~\ref{fig:framework} presents the overview of our \method framework. First it learns a constrained multi-tree based model with the training data. By adopting every prefix path from the root of a tree to any of its non-leaf nodes as a discriminative pattern, a large pool of discriminative patterns is ready for further top-$k$ discriminative pattern selection. Two different solutions, forward selection and LASSO, are utilized to select top-$k$ discriminative patterns based on their performances using a generalized linear model. Both solutions have shown high accuracies in the experiments. The corresponding linear model with the selected top-$k$ discriminative patterns is adopted to make predictions on new examples. Our \method is extremely fast and memory-efficient.

    \vspace{-0.1in}
    \subsection{Discriminative Pattern Generation}
        The first component in the \method framework is the generation of high-quality discriminative patterns, as shown in Algorithm~\ref{alg:dpclass-1}. We use \textit{tree bag} to refer the set of instances falling into a specific node in the decision tree. The random decision tree~\cite{chen2004using} introduces the randomness via bootstrapping training data, randomly selecting features and splitting values when dividing a large tree bag into two smaller ones. $T$ random decision trees are generated, and for each tree, all prefix paths from its root to non-leaf nodes are treated as discriminative patterns. Due to the predictivity of decision trees, so-generated patterns are highly effective in the specific prediction task. Note that the decision tree is built with different loss functions in different tasks, which could be entropy gain in classification tasks or the mean square error in regression tasks.

        \SetAlgoSkip{}
        \begin{algorithm}[htbp]
            \caption{Discriminative Pattern Generation}\label{alg:dpclass-1}
            \textbf{Require}: $n$ training instances $(\mathbf{x}_i, y_i)$, the number of trees $T$, the depth threshold $D$, and minimum tree bag size $\sigma$ \\
            \textbf{Return}: a set of discriminative patterns for further selection. \\
            $\mathcal{P} \leftarrow \emptyset$ \\
            \For {$t = 1$ {\bfseries to} $T$} {
                Build a random decision tree~\cite{chen2004using} with maximum depth $D$ and minimum tree bag size $\sigma$. \\
                \For {each non-leaf node $u$} {
                    $\mathcal{P} \leftarrow \mathcal{P} \cup \{root \rightarrow u\}$\\
                }
            }
            \textbf{return} $\mathcal{P}$
        \end{algorithm}

        In real-world datasets, the discriminative patterns are frequently emerging, and the length of such patterns are not too long. Specifically, we assume that the number of instances satisfying a given discriminative pattern should be at least $\sigma$, and the length of discriminative patterns is no more than $D$. The returned patterns are discriminative to ensure prediction accuracy and diverse to ensure sufficient condition coverage. As one of the most famous multi-tree based models, random forest~\cite{chen2004using} is the best fit addressing all the requirements if we treat every prefix path from the root of a tree to its non-leaf node as a discriminative pattern. First, distributions of labels of instances in a tree bag always have low entropy. Therefore, the patterns are discriminative on the training data. Second, it provides many putative patterns from various random decision trees trained on different bootstrapped datasets. Third, the depth threshold $D$ and the minimum tree bag size $\sigma$ can be naturally added as constraints during the growth of trees.

    \vspace{-0.1in}
    \subsection{Pattern Space Construction}
        After the pattern generation, \method maps the instances in the original feature space to a new pattern space using the set of discriminative patterns discovered by tree models, as shown in Algorithm~\ref{alg:dpclass-1.5}. For each discriminative pattern, there is one corresponding binary dimension describing whether the instances satisfy the pattern or not. Because the dimension of the pattern space is equal to the number of discriminative patterns which is a very large number after the generation phase, we need to further select a limited number of patterns and thus make the pattern space small and efficient. It is also worth a mention that this mapping process is able to be fully parallelized for speedup.

        \SetAlgoSkip{}
        \begin{algorithm}[htbp]
            \caption{Pattern Space Construction}\label{alg:dpclass-1.5}
            \textbf{Require}: $n$ instances $(\mathbf{x}_i)$, a discriminative patterns set $\mathcal{P}$\\
            \textbf{Return}: $n$ instances in pattern space $(\mathbf{x'}_i)$ \\
            \For{$i = 1$ {\bfseries to} $n$} {
                $\mathbf{x'}_i \leftarrow \mathbf{0}$ \\
                \For{$j$-th pattern $P_j$ in $\mathcal{P}$} {
                    \If {$\mathbf{x}_i$ satisfies pattern $P_j$} {
                        $\mathbf{x'}_{i,j} \leftarrow 1$ \\
                    }
                }
            }
            \textbf{return} $(\mathbf{x'}_i)$
        \end{algorithm}

    \vspace{-0.1in}
    \subsection{Top-k Pattern Selection}

        After a large pool of discriminative patterns is generated, further top-$k$ selection needs to be done to identify the most informative and interpretable patterns. A naive way is to use heuristic functions, such as information gain and gini index, to evaluate the significance of different patterns on the prediction task and choose the top ranked patterns. However, the effects of top ranked patterns based on the simple heuristic scores may have a large portion of overlaps and thus their combination does not work optimally. Therefore, to achieve the best performance and find complementary patterns, we propose two effective solutions: forward selection and LASSO, which make decisions based on the effects of the pattern combinations instead of considering different patterns independently.

    \vspace{-0.1in}
    \subsubsection{Forward Pattern Selection}
        Instead of exhausted search of all possible combinations of $k$ discriminative patterns, forward selection gradually adds the discriminative patterns one by one while each newly added discriminative pattern is the best choice at that time~\cite{derksen1992backward}, which provides an efficient approximation of the exhausted search. To be more specific, when the first $k'$ discriminative patterns are fixed, the algorithm empirically adds one more discriminative pattern so that the new set of $k'+1$ patterns achieves the best training performance in the generalized linear model, as shown in Algorithm~\ref{alg:dpclass-2.1}. As mentioned before, when assuming training and testing data have the same distribution, using training accuracy is very reasonable.

        \SetAlgoSkip{}
        \begin{algorithm}[htbp]
            \caption{Top-$k$ Pattern Selection: Forward}\label{alg:dpclass-2.1}
            \textbf{Require}: $n$ training examples $(\mathbf{x}_i, y_i)$, a set of discriminative patterns $\mathcal{P}$ and $k$ \\
            \textbf{Return}: top-$k$ discriminative patterns set $\mathcal{P}_k$ and a generalized linear model $f(\cdot)$ \\
            $\mathcal{P}_k \leftarrow \emptyset$ \\
            \For {$t = 1$ {\bfseries to} $k$} {
                \For {each pattern $p$ in $\mathcal{P}$} {
                    $\mathbf{x'} \leftarrow $ construct pattern space($\mathbf{x}, \mathcal{P}_k \cup \{p\}$) using Algorithm~\ref{alg:dpclass-1.5} \\
                    $g(\cdot) \leftarrow$ a generalized linear model~\cite{suykens1999least} on $(\mathbf{x'}_i, y_i)$\\
                    $per_p \leftarrow g(\cdot)$'s training performance
                }
                $\mathcal{P}_k \leftarrow \mathcal{P}_k \cup \{\arg\max_{p} per_p\}$ \\
            }
            $\mathbf{x'} \leftarrow $ construct pattern space($\mathbf{x}, \mathcal{P}_k$) \\
            $f(\cdot) \leftarrow$ a generalized linear model on $(\mathbf{x'}_i, y_i)$\\
            \textbf{return} $\mathcal{P}_k$, $f(\cdot)$
        \end{algorithm}

    \vspace{-0.1in}
    \subsubsection{LASSO based Pattern Selection}
        L1 regularization (i.e., LASSO~\cite{tibshirani1996regression}) is designed to make the weight vector sparse by tuning a nonnegative parameter $\lambda$, where the features with non-zero weight will be the selected ones. Since we are actually selecting features in the pattern space, for a given $\lambda$, we optimize the following loss function to get a subset of important patterns.
        \begin{equation}
            \label{eq:lasso}
            \mathcal{L} = \sum_{i}^{n} l(\mathbf{x'}_i^T \mathbf{w}, y_i) + \lambda \cdot \|w\|_1,
        \end{equation}
        where $\mathbf{x'}_i$ is the mapped binary feature representation in pattern space of $i$-th example; $\mathbf{w}$ is the weight vector in the generalized linear model; $l(\cdot, \cdot)$ is a general loss function such as logistic loss. To ensure there are at most $k$ patterns having non-zero weights in the pattern space, we should carefully choose a value for $\lambda$. We assume that there exists hidden importance among the features. When the weight of a feature is non-zero in a given $\lambda = v$, it is also non-zero for any smaller $\lambda < v$. A binary search algorithm is shown in Algorithm~\ref{alg:dpclass-2.2}. The LASSO implementation in GLMNET~\cite{friedman2009glmnet} is adopted in this thesis, whose loss function is the cross entropy.

        \SetAlgoSkip{}
        \begin{algorithm}[htbp]
            \caption{Top-$k$ Pattern Selection: LASSO}\label{alg:dpclass-2.2}
            \textbf{Require}: $n$ training examples $(\mathbf{x}_i, y_i)$, a set of discriminative patterns $\mathcal{P}$, $k$, and a small value $\epsilon$\\
            \textbf{Return}: top-$k$ discriminative patterns $P_i$ and a generalized linear model $f(\cdot)$ \\
            $\mathcal{P}_k \leftarrow \emptyset$ \\
            $l \leftarrow 0$, $r \leftarrow +\infty$ \\
            $\mathbf{x'} \leftarrow $ construct pattern space($\mathbf{x}, \mathcal{P}$) using Algorithm~\ref{alg:dpclass-1.5} \\
            \While{$l + \epsilon < r$} {
                $\lambda \leftarrow (l + r) / 2$ \\
                $\mathbf{w} \leftarrow \arg \min_{\mathbf{w}} $ Equation~\ref{eq:lasso}\\
                \If {non-zero weighted patterns $\le k$} {
                    $\mathcal{P}_k \leftarrow \{p | p \mbox{'s weight is non-zero}\}$ \\
                    $r \leftarrow \lambda$
                } \Else {
                    $l \leftarrow \lambda$
                }
            }
            $\mathbf{x'} \leftarrow $ construct pattern space($\mathbf{x}, \mathcal{P}_k$) \\
            $f(\cdot) \leftarrow$ a generalized linear model on $(\mathbf{x'}_i, y_i)$\\
            \textbf{return} $\mathcal{P}_k$,  $f(\cdot)$
        \end{algorithm}

        \SetAlgoSkip{}
        \begin{algorithm}[htbp]
            \caption{Prediction}\label{alg:dpclass-3}
            \textbf{Require}: $n$ testing examples $(\mathbf{x}_i)$, top-$k$ discriminative patterns set $\mathcal{P}_k$, and the generalized linear model $f(\cdot)$\\
            \textbf{Return}: predictions of testing instances $\hat{y}_i$ \\
            $\mathbf{x'} \leftarrow $ construct pattern space($\mathbf{x}, \mathcal{P}_k$) using Algorithm~\ref{alg:dpclass-1.5} \\
            \For{$i = 1$ {\bfseries to} $n$} {
                $\hat{y}_i \leftarrow f(\mathbf{x'}_{i})$ \\
            }
            \textbf{return} $\hat{y}$
        \end{algorithm}

    \vspace{-0.1in}
    \subsection{Prediction}
        Once the top-$k$ discriminative patterns are determined, for any upcoming new test instance, \method first maps it into the learned pattern space, and then applies the pre-trained generalized linear model to compute the prediction, as shown in Algorithm~\ref{alg:dpclass-3}. As the number of patterns is limited, both the mapping into the pattern space and the prediction of the generalized linear model will be extremely fast.

    \vspace{-0.1in}
    \subsection{Time Complexity Analysis}
        To build up a single random decision tree with depth threshold $D$ and minimum tree bag size $\sigma$, by assuming both numbers of random features and random partitions are small and fixed constants, the time complexity is $O(nD)$, because the total number of instances on each level of the tree is $n$. Therefore, the time complexity of generating $T$ trees is $O(TnD)$ in the generation step.

        For the selection step, the complexity is mainly determined by the number of discriminative patterns induced by $T$ random decision trees, which is dependent on the total number of non-leaf nodes. As the maximum depth of a single tree is $D$, there is an upper bound on number of leaf nodes $2^D$. Starting from the tree bag size, the number of leaf nodes should be no more than $\lceil \frac{n}{\sigma} \rceil$. Since the trees here are all binary trees, the number of leaf nodes is one more than the number of non-leaf nodes. Therefore, the number of discriminative patterns $|\mathcal{P}|$ (i.e., the number of non-leaf nodes) is bounded by $T \cdot \min\{2^D, \lceil \frac{n}{\sigma} \rceil\} - 1$. If we solve logistic regression and LASSO using (sub-)gradient descent algorithm, and thus the time complexity per gradient step is only linear to the dimension of features and the number of examples. The time complexity is proportional to $O(|\mathcal{P}| \cdot n \cdot k^2)$ if forward selection is used, while it is proportional to $O(n \cdot k \cdot |\mathcal{P}|)$ if LASSO is used. By assuming the numbers of iterations to converge are similar in LASSO and forward selection, LASSO will be a little more efficient than forward selection.

        When predicting new test instances, one can easily figure out the bottleneck is mapping instances into the learned pattern space. Therefore, in the batch mode where examples are considered together, the time complexity is $O(n \cdot k \cdot D)$.  In the streaming (or online) mode where instances come one by one, the time complexity is $O(k \cdot D)$, where $k$ is the number of discriminative patterns and $D$ is the maximum tree depth, which is equivalent to the maximum number of conditions in a single pattern.

        It is worth mentioning that all modules can be fully parallelized, leading to further speedup in practice. 
\section{Experiments}

    In this section, we conduct extensive experiments to demonstrate the interpretability, efficiency and effectiveness of our proposed \method framework. We first introduce our experimental settings, discuss the efficiency and interpretability, and then give the results on classification and regression tasks as well as parameter analysis.

    \vspace{-0.1in}
    \subsection{Experimental Settings}

        This subsection presents the datasets, baseline methods, and learning tasks in our experiments.

        \begin{table}[t]
            \caption{The statistics of our real-world datasets from UCI Machine Learning Repository for classification and regression.} \label{tbl:dataset}
            \center
        \vspace{-0.2in}
        \scalebox{0.85}{
            \begin{tabular}{|l|l|r|r|l|}
            \hline
            Type & Dataset & \# Instances & \# Dimensions & Variable type \\
            \hline \hline
            \multirow{6}{*}{Classification} & Adult & 45,222 & 14 & Mixed \\
            \cline{2-5}
            & Hypo & 3,772 & 19 & Mixed\\
            \cline{2-5}
            & Sick & 3,772 & 19 & Mixed \\
            \cline{2-5}
            & Chess & 28,056 & 6 & Mixed\\
            \cline{2-5}
            & Crx & 690 & 15 & Mixed\\
            \cline{2-5}
            & Sonar & 208 & 60 & Numeric\\
            \hline
            \hline
            \multirow{3}{*}{High dimension} & Nomao & 29,104 & 120 & Mixed \\
            \cline{2-5}
            & Musk & 7,074 & 166 & Numeric \\
            \cline{2-5}
            & Madelon & 1,300 & 500 & Numeric\\
            \hline
            \hline
            \multirow{3}{*}{Regression} & Bike & 17,379 & 10 & Mixed \\
            \cline{2-5}
            & Parkinsons & 5,875 & 16 & Numeric \\
            \cline{2-5}
            & Crime & 1,994 & 99 & Numeric \\
            \hline
            \end{tabular}
        }
        \vspace{-0.2in}
        \end{table}

        \vspace{-0.1in}
        \subsubsection{Datasets}

            First, we generate synthetic datasets where the features are demographics and lab test results of patients and the label is whether the patient has a disease, in order to demonstrate the interpretability of \method. Assuming doctors can diagnose the disease using some rules based on these information, it can be verified whether the top discriminative patterns selected by \method are consistent with the actual diagnosing rules.

            Several real world classification and regression datasets from UCI Machine Learning Repository are used in the experiments, as shown in Table~\ref{tbl:dataset} with statistics of the number of instances and the number of features. In the datasets \textit{adult}, \textit{hypo} and \textit{sick}, the ratio of standard train/test splitting is $2:1$. Therefore, for the other classification and regression datasets, we divide the datasets into train/test ($2:1$) by unbiased sampling as preprocessing.

            For classification tasks, to compare with \ddpmine, we use the same datasets including \textit{adult}, \textit{hypo}, \textit{sick}, \textit{crx}, \textit{sonar}, \textit{chess}, \textit{waveform}, and \textit{mushroom}. Because both \ddpmine and \method achieve almost perfect accuracy (very close to $100\%$) on the datasets \textit{waveform} and \textit{mushroom}, these two datasets are omitted. In addition, the performance of \method on high-dimensional datasets (\textit{nomao}, \textit{musk} and \textit{madelon} datasets) is also investigated, since \ddpmine performs poorly on high-dimensional data. The metric is the accuracy on the testing data: higher accuracy means better performance.

            For regression datasets, we choose general datasets such as \textit{bike} and \textit{crime}, as well as clinical datasets where patterns are more likely to be present, such as \textit{parkinsons}. Furthermore, to make the errors in different datasets comparable, min-max normalization is adopted to scale the continuous labels into $[0, 1]$. The metric is the rooted mean square error (RMSE) on the testing data: a lower RMSE means better performance.

            \begin{table}[t]
                \center
                \caption{Model complexity and computational complexity. Model complexity is measured by the number of encoded patterns. Here $D$ is the number of dimensions, $k$ is the number of top patterns, and $T$ is the number of trees.} \label{tbl:model size}
                \begin{tabular}{|l|l|l|}
                \hline
                Model & Model complexity (\# Patterns) & Time complexity \\
                \hline
                \hline
                \method & $k \approx 20\sim50$ & $O(k \cdot D)$ \\
                \hline
                DT & \# of nodes $\approx 64$ & $O(D)$ \\
                \hline
                DDPMine & $k \approx 100\sim1,000$ & $O(k \cdot D)$\\
                \hline
                LRF & \# of nodes $\approx 6,400$ & $O(T \cdot D)$ \\
                \hline
                RF & \# of nodes $\ge 10,000$ & $O(T \cdot D)$ \\
                \hline
                \end{tabular}
                \vspace{-0.2in}
            \end{table}

        \vspace{-0.1in}
        \subsubsection{Baseline Methods}

            \ddpmine \cite{cheng2008direct} is a previous state-of-the-art discriminative pattern based algorithm. It first discretizes the continuous variables such that frequent pattern mining algorithm could be applied. Using frequent and discriminative patterns, new feature space is constructed and any classical classifiers could be further utilized. \ddpmine only focuses on classification tasks and it is not applicable in regression experiments.

            Random Forest (\textsc{RF}) \cite{chen2004using} is another baseline method using same parameters as those in the random forest used in \method, except for $D$. There is no limit on the depth in \textsc{RF}. Moreover, we are interested in the limited-depth random forest model (\textsc{LRF}) built in the top-$k$ generation step of global patterns. These two tree-based methods are capable in both classification and regression tasks. It is expected if these two complex models (i.e., hard to interpret) have slightly better performance than \method, because the major contributions of \method are the concise interpretable patterns instead of solely the accuracy. To make a fair comparison, Decision Tree (\textsc{DT}) with a similar number of nodes with \method is also listed as a baseline.

        \vspace{-0.1in}
        \subsubsection{Classification and Regression Tasks}
            In \method, for the classification tasks, the default parameter setting is $T = 100, D = 6, \sigma = 10, k = 20$. For the regression tasks, because the continuous labels are more complex than those discrete class labels in classification, it is natural to incorporate more patterns. Therefore, the default setting is $T = 100, D = 6, \sigma = 10, k = 30$.

            We will show results using both forward selection (\method-F) and LASSO (\method-L) to select the top-$k$ discriminative patterns. We deeply study the impact of the parameters such as the number of selected discriminative patterns $k$ and the number of trees in the random forest $T$. Therefore, we fix the other parameters as their default values and vary the parameter value to study their impacts, respectively.

    \vspace{-0.1in}
    \subsection{Efficiency and Interpretability}

        \textit{Efficiency.} The test running time is linearly proportional to the model complexity, which is related to the number of patterns the model used. In the experiments, \ddpmine needs 100 to 1,000 patterns while \method only needs 20, which indicates a significant reduction of prediction runtime. Moreover, the random forest without any constraints will contain more than 10,000 nodes (i.e., patterns), which is far more expensive. Although the evaluation of random forest for a single testing instance will traverse only a number of nodes equals to the sum of depths in different trees, it always needs more than 1,000 traverses in the experiments. Therefore, \method is the most efficient model for testing new instances, compared to \ddpmine and random forest, by achieving about \textit{20 to 50 times speedup} in practice. Furthermore, \method could be fully parallelized for further speedup. The empirical results are presented in Table~\ref{tbl:model size}.

        \textit{Interpretability: our discovery of interpretable patterns.}
        We generate a small medical dataset for binary classification to demonstrate the interpretability. For each patient, we draw several uniformly sampled features as follows:
        \begin{enumerate}
            \item Age (A): positive integers no more than 60.
            \item Gender (G): male or female.
            \item Lab Test 1 (LT1): blood types (categorical values) from \{A, B, O, AB\}.
            \item Lab Test 2 (LT2): continuous values in $[0,1]$.
        \end{enumerate}
        Totally, there are $10^5$ random patients for training and $5 \cdot 10^4$ patients for testing.

        The positive label of the disease is assigned to a patient if at least one of the following rules holds:
        \begin{enumerate}
            \item (A $>$ 18) and (G $=$ Male) and (LT1 $=$ AB) and (LT2 $\ge$ 0.6),
            \item (A $>$ 18) and (G $=$ Female) and (LT1 $=$ O) and (LT2 $\ge$ 0.5),
            \item (A $\le$ 18) and (LT2 $\ge$ 0.9).
        \end{enumerate}

        To make the classification tasks more challenging, $0.1\%$ noise is added to the training data. That is, $0.1\%$ labels in training will be flipped.

        We apply both \method-F and \method-L on this dataset. Both give the test accuracy $99.99\%$. The top-3 discriminative patterns found in both \method-F and \method-L are listed as below. We observe that the found patterns are quite close to the groundtruth rules. We demonstrate that the selected discriminative patterns provide high-quality explanation:
        \begin{enumerate}
            \item (A $>$ 18) and (G $=$ Female) and (LT1 $=$ O) and (LT2 $\ge$ 0.496),
            \item (A $\le$ 18) and (LT2 $\ge$ 0.900),
            \item (A $>$ 18) and (G $=$ Male) and (LT1 $=$ AB) and (LT2 $\ge$ 0.601).
        \end{enumerate}

        We apply \ddpmine to this dataset but its accuracy is only $95.64\%$, because the discretization brings too much noise. The top-3 patterns mined by \ddpmine are as follows, which are quite different from expectation:
        \begin{enumerate}
            \item (LT2 $>$ 0.8),
            \item (G $=$ Male) and (LT1 $=$ AB) and (LT2 $\ge$ 0.6) and (LT2 $<$ 0.8),
            \item (G $=$ Female) and (LT1 $=$ O) and (LT2 $\ge$ 0.6) and (LT2 $<$ 0.8).
        \end{enumerate}

    \vspace{-0.1in}
    \subsection{Effectiveness in Classification}

        \ddpmine is a previous state-of-the-art pattern-based classification method, which outperforms traditional classification models including decision tree and support vector machine \cite{cheng2007discriminative}\cite{cheng2008direct}. We compare \method, \ddpmine and \textsc{RF} on the same datasets used in \ddpmine. The results are shown in Table~\ref{tbl:accuracy}.
        \method-F and \method-L always have higher accuracy over \ddpmine. An important reason of this advantage is that the candidate patterns generated by tree-based models in \method are much more discriminative and thus more effective on the specific classification task than those frequent but less useful patterns extracted in \ddpmine. Except for \textit{sick} dataset, \method-F has the highest accuracy, while \method-L works best on sick dataset. It seems that \method-F works a little better than \method-L. However, their results are quite close to each other and are both better than those of \ddpmine on most datasets.

        More surprisingly, \method demonstrates even better performance than the complex model random forest on several datasets, while its accuracies on other datasets are still comparable with \textsc{RF}, which is due to the effectiveness of the pattern selection module where we select the optimal pattern combination instead of selecting patterns independently. This shows that the proposed model is very effective in classification tasks while it is highly concise and interpretable.

        \begin{table}[t]
            \center
            \caption{Test Accuracy on Classification Datasets tested in \ddpmine. \ddpmine outperforms decision tree and support vector machine on all these datasets~\cite{cheng2007discriminative, cheng2008direct}. \method can achieve the best performance in almost every dataset, while RF is the best on the chess dataset.} \label{tbl:accuracy}
        \scalebox{0.9}{
            \begin{tabular}{|l||r|r|r|r|r|r|}
            \hline
            Dataset & adult & hypo & sick & crx & sonar & chess \\
            \hline \hline
            \method-F & \textbf{85.66\%} & \textbf{99.58\%} & 98.35\% & \textbf{89.35\%} & \textbf{85.29\%} & 92.25\%  \\
            \hline
            \method-L & 84.33\%  &  99.28\% & \textbf{98.87\%} & 87.96\% & 83.82\% & 92.05\%  \\
            \hline
            \textsc{DT} & 83.33\% & 92.90\% & 93.82\% & 77.78\% & 67.65\% & 89.86\%  \\
            \hline
            \ddpmine & 83.42\% & 92.69\% & 93.82\% & 87.96\% & 73.53\% & 90.04\%  \\
            \hline
            \textsc{LRF} & 83.51\% & 95.78\% & 93.93\% & \textbf{89.35\%} & 83.82\% & 90.04\%  \\
            \hline
            \textsc{RF} & 85.45\% & 97.22\% & 94.03\% & \textbf{89.35\%} & 83.82\% & \textbf{94.22\%}  \\
            \hline
            \end{tabular}
        }
        \vspace{-0.5cm}
        \end{table}

        \begin{table}[t]
            \center
            \caption{Testing RMSE on regression tasks. To make the errors in different datasets comparable, min-max normalization is adopted to scale the continuous labels into $[0, 1]$. Our \method methods take much fewer patterns than \textsc{RF} and perform significantly better than the simple \textsc{DT} and \textsc{LRF} models.} \label{tbl:reg}
            \begin{tabular}{|l||r|r|r|r|r|}
            \hline
            Dataset & bike & crime & parkinsons & Diff  \\
            \hline \hline
            \method-F & 0.0872 & 0.1515 & 0.1969 &  N/A  \\
            \hline
            \method-L & 0.0974 & 0.1465 & 0.1951 &  N/A  \\
            \hline
            \textsc{DT} & 0.1186 & 0.1971 & 0.2129 & +24.74\% \\
            \hline
            \textsc{LRF} & 0.1211 & 0.1367 & 0.1976 & +16.64\% \\
            \hline
            \textsc{RF} & \textbf{0.0836} & 0.\textbf{1372} & \textbf{0.1865} & - 6.77\%   \\
            \hline
            \end{tabular}
            \vspace{-0.2cm}
        \end{table}

        \begin{table}[t]
            \center
            \caption{Testing accuracy on high dimensional datasets. \method performs consistently better than \ddpmine, and it is comparable with the complex \textsc{RF} and better on \textit{madelon}.} \label{tbl:highdim}
            \begin{tabular}{|l||r|r|r|}
            \hline
            Dataset & nomao & musk & madelon\\
            \hline \hline
            \method-F  & 97.17\% & 95.92\% & 74.50\% \\
            \hline
            \method-L &  96.94\% & 95.71\% & \textbf{76.00\%} \\
            \hline
            \textsc{DT} &  92.98\% & 87.82\% & 50.34\% \\
            \hline
            \ddpmine &  96.83\% & 93.29\% & 59.83\% \\
            \hline
            \textsc{LRF} &  95.56\% & 90.49\% & 59.17\% \\
            \hline
            \textsc{RF} &  \textbf{97.86\%} & \textbf{96.60\%} & 56.50\% \\
            \hline
            \end{tabular}
\vspace{-0.5cm}
        \end{table}

        \begin{figure*}[t]
          \centering
          \subfigure[Classification: \method-F] {
            \includegraphics[width=0.23\textwidth]{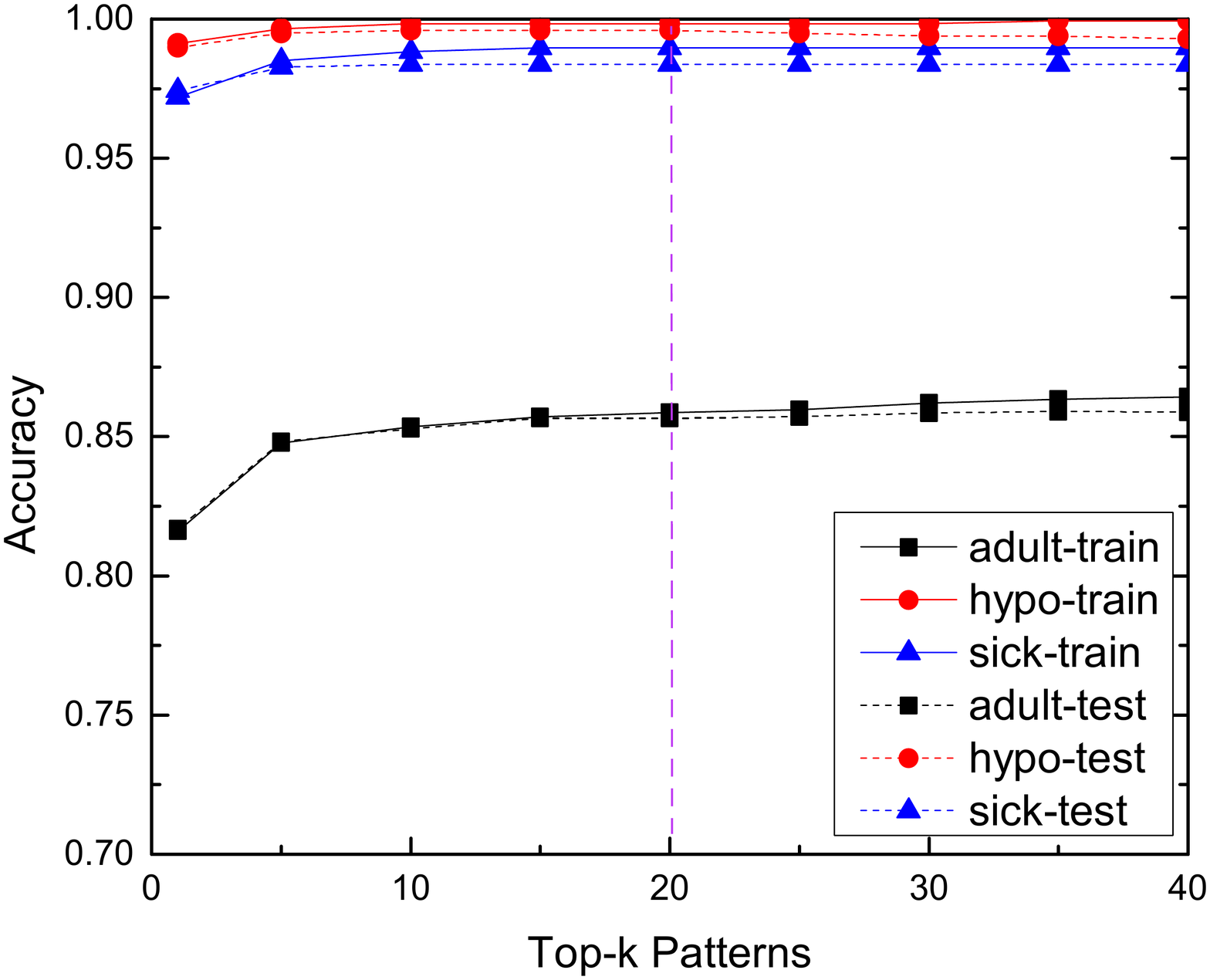}
          }
          \subfigure[Classification: \method-L] {
            \includegraphics[width=0.23\textwidth]{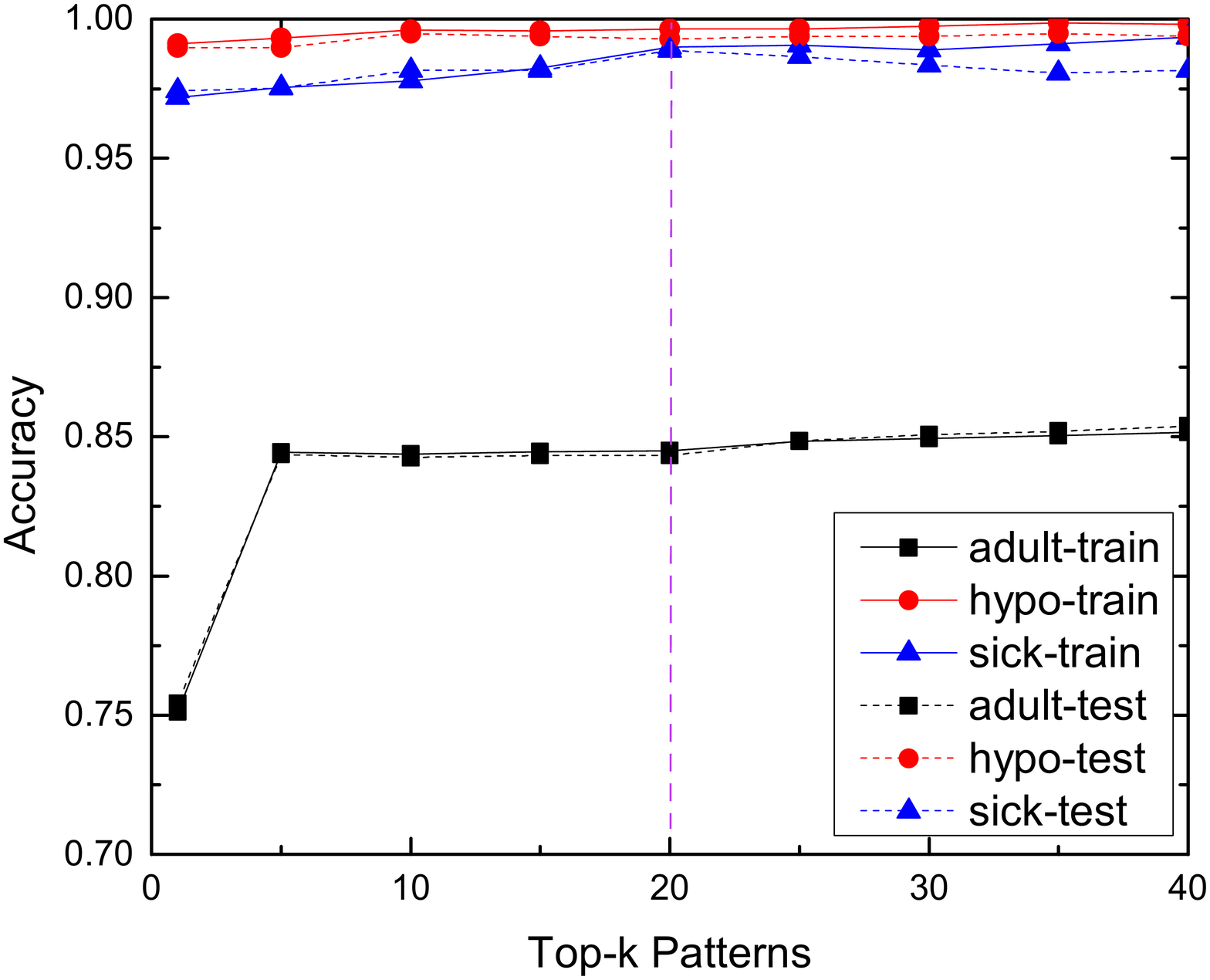}
          }
          \subfigure[Regression: \method-F] {
            \includegraphics[width=0.23\textwidth]{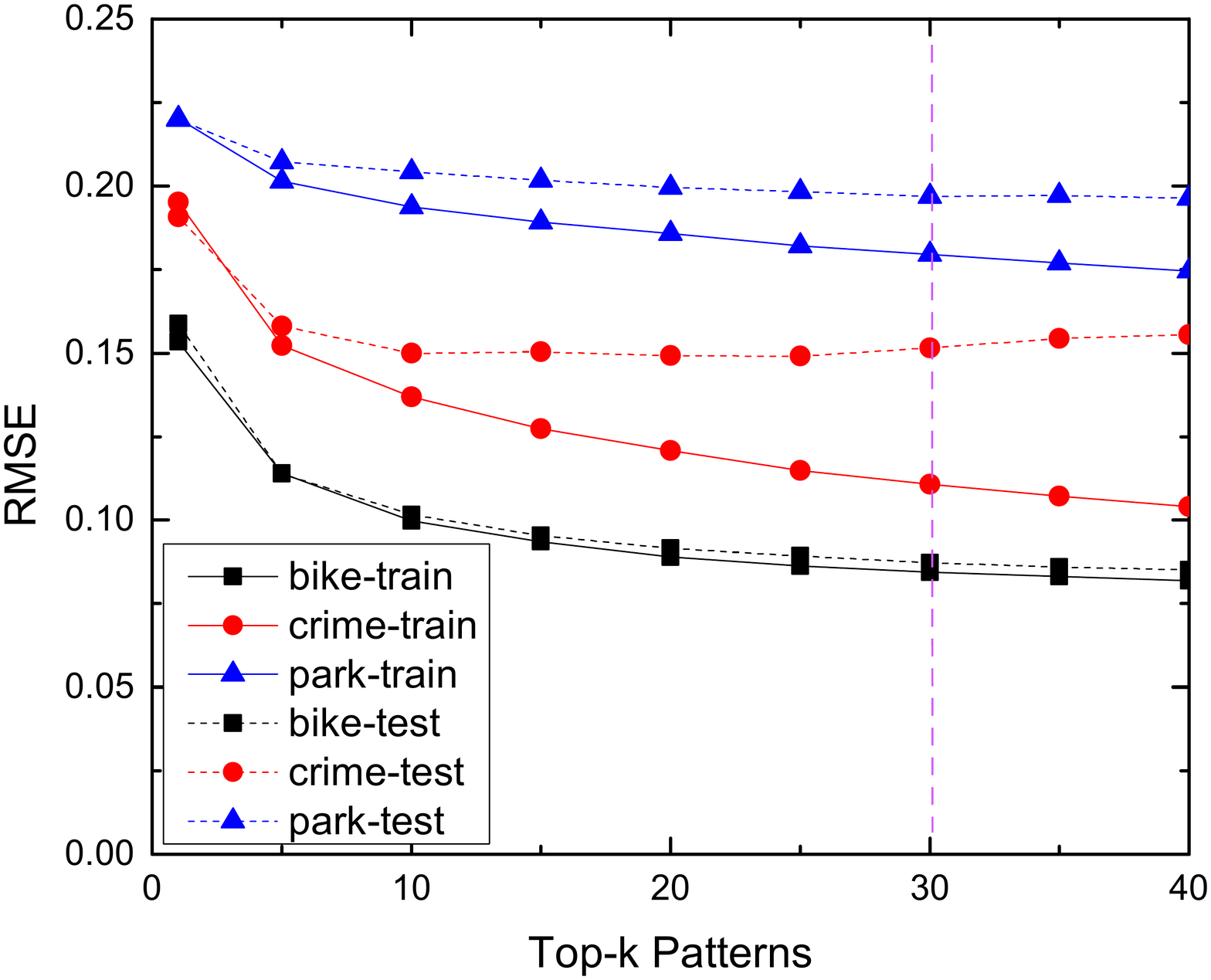}
          }
          \subfigure[Regression: \method-L] {
            \includegraphics[width=0.23\textwidth]{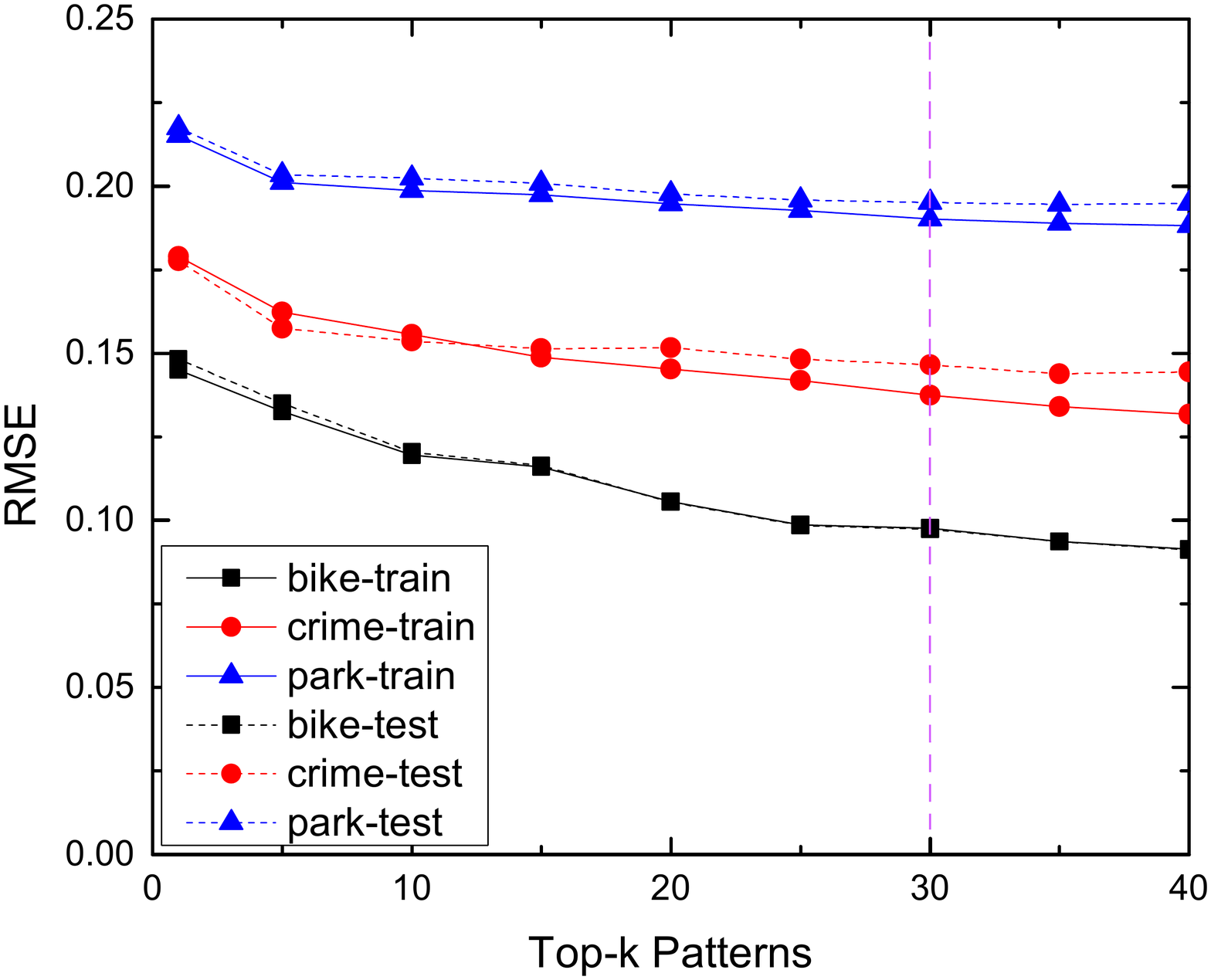}
          }
          \vspace{-0.1in}
          \caption{The impact of top-$k$ patterns in classification and regression tasks. Training and testing performances are almost overlapped in some datasets. We observe that a small number of patterns (e.g., 20 for classification and 30 for regression) are enough to achieve stable performance.}
          \label{fig:vary_k}
\vspace{-0.5cm}
        \end{figure*}

    \vspace{-0.1in}
    \subsection{Effectiveness in Regression}

        Since \ddpmine is not applicable on regression tasks, we only compare \method with \textsc{DT}, \textsc{RF}, and \textsc{LRF}. Note that these two methods are highly complicated and thus preserve very limited interpretability. The RMSE results and the average differences compared to \method are shown in Table~\ref{tbl:reg}.

        Unlike the results in classification datasets, complex models outperform \method on all datasets although the difference is not very significant. This is reasonable because, different from the discrete class labels, the real valued prediction increases the level of difficulty. Although we have raised the number of top patterns a little, bag-of-patterns feature representations based on a small number of patterns still have some limitations to predict a real value. For example, there are at most $2^{30}$ different examples in the constructed pattern space, which means there are at most $2^{30}$ different predicted values, but infinite real numbers are likely to be the true value for a new example. However, it is worth noting that \method (both \method-F and \method-L) always achieves comparable performance with \textsc{RF}, and work better than or similar to \textsc{DT} and \textsc{LRF}, which still demonstrates the effectiveness of \method to some extent while the model is more compact and interpretable than \textsc{RF} and \textsc{LRF}.

    \vspace{-0.1in}
    \subsection{Effectiveness in High Dimensions}

        We are interested in high-dimensional datasets (i.e., at least 100 dimensions) because \ddpmine is not effective in large dimensional data. To compare with \ddpmine, we use classification datasets whose number of dimensions is at least 100 and no regression datasets are used.  As the dimension of the original feature space grows, it is reasonable to increase the depth threshold $D$, as well as the number of trees $T$, to involve higher order interactions and increase the number of candidate discriminative patterns. Therefore, we set $D = 10$ and $T = 200$. Meanwhile, the dimension of mapped pattern space may also need to be increased due to the higher complexity of problems. As a result, we set $k = 50$ in \textit{nomao} and \textit{musk} datasets. However, we kept $k=20$ in \textit{madelon} dataset because many features are noises.

        As shown in Table~\ref{tbl:highdim}, \method can always outperform \ddpmine and generate comparable results to those by \textsc{RF}. It is worth noting that in \textit{madelon} dataset, \method-F and \method-L outperform \textsc{RF} significantly. As stated before, \textit{madelon} is highly noisy. As a result, many patterns generated by random forest are not that reliable, which can be very poor at test data although they are discriminative in training data. On the other hand, \method compresses the patterns and only keeps the most discriminative ones, and thus alleviates this problem to some extent. This demonstrates the robustness of \method especially when the features are high dimensional and noisy. It is also worth a mention that the training process of \method is at least 10 times faster than \ddpmine in high dimensional datasets.

    \vspace{-0.1in}
    \subsection{Parameter Analysis}

        In this subsection, we deeply study the parameters including the number of top patterns $k$ and the number of trees in the random forest $T$.

        \begin{figure*}[t]
          \centering
          \subfigure[Classification: \method-F] {
            \includegraphics[width=0.23\textwidth]{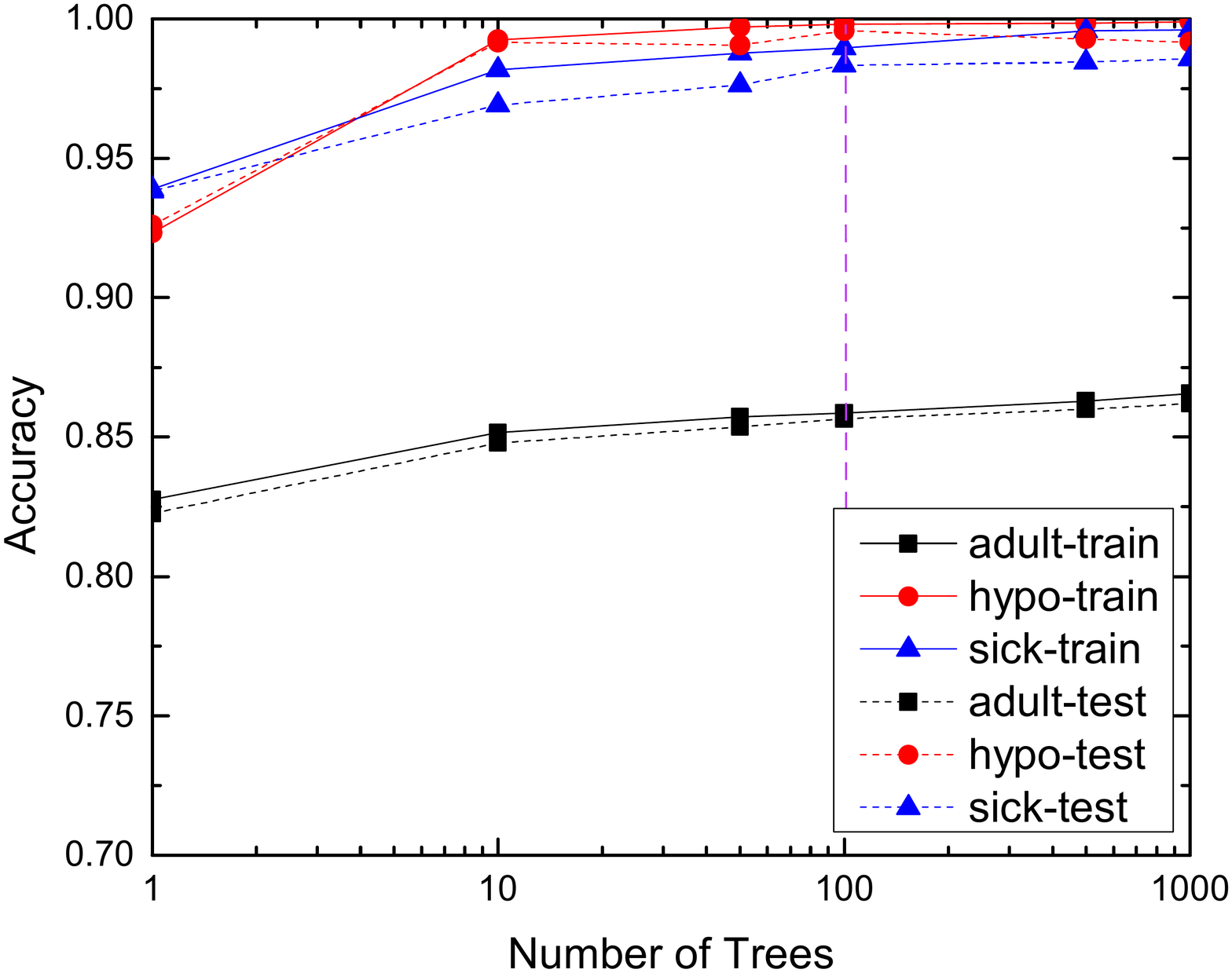}
          }
          \subfigure[Classification: \method-L] {
            \includegraphics[width=0.23\textwidth]{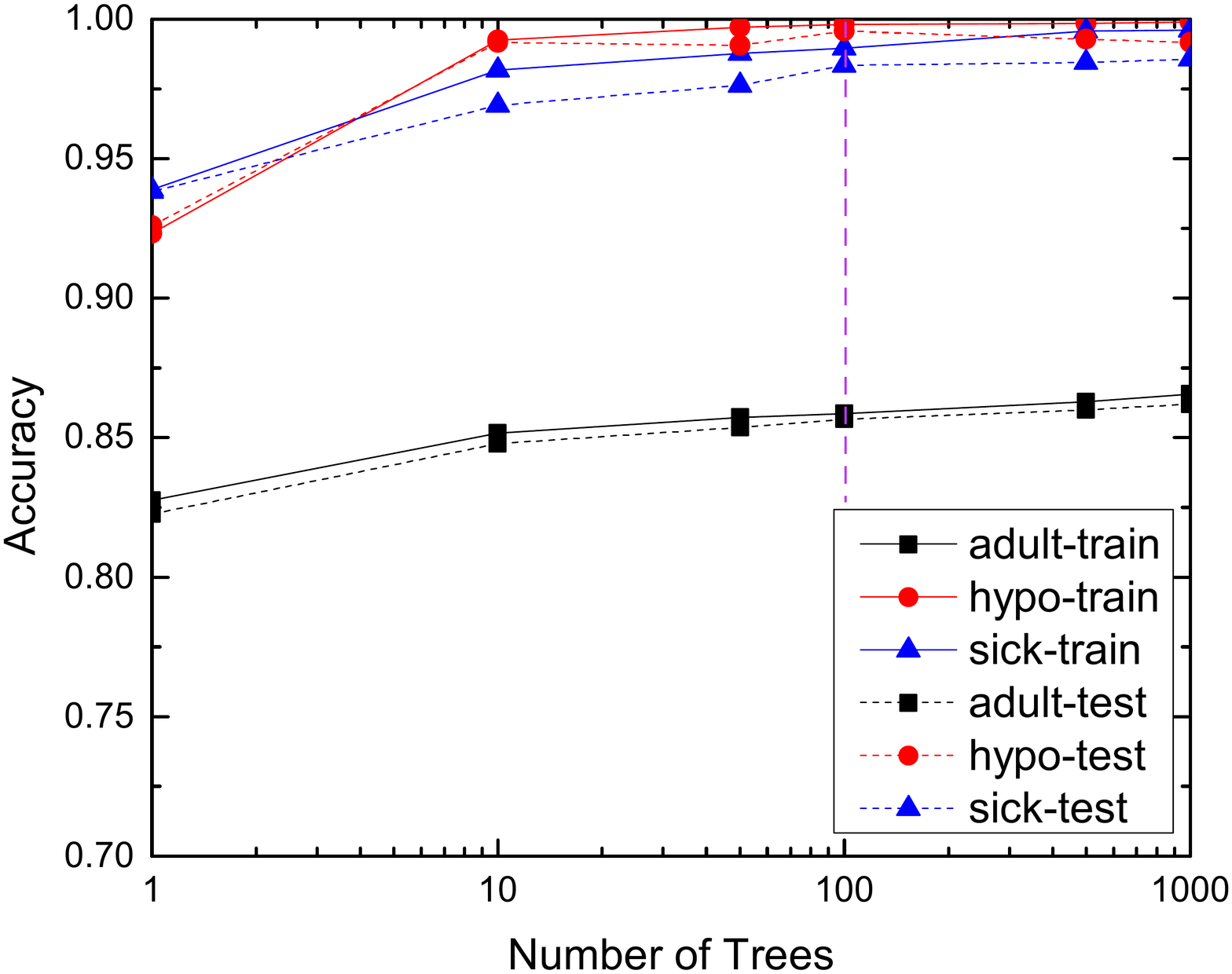}
          }
          \subfigure[Regression: \method-F] {
            \includegraphics[width=0.23\textwidth]{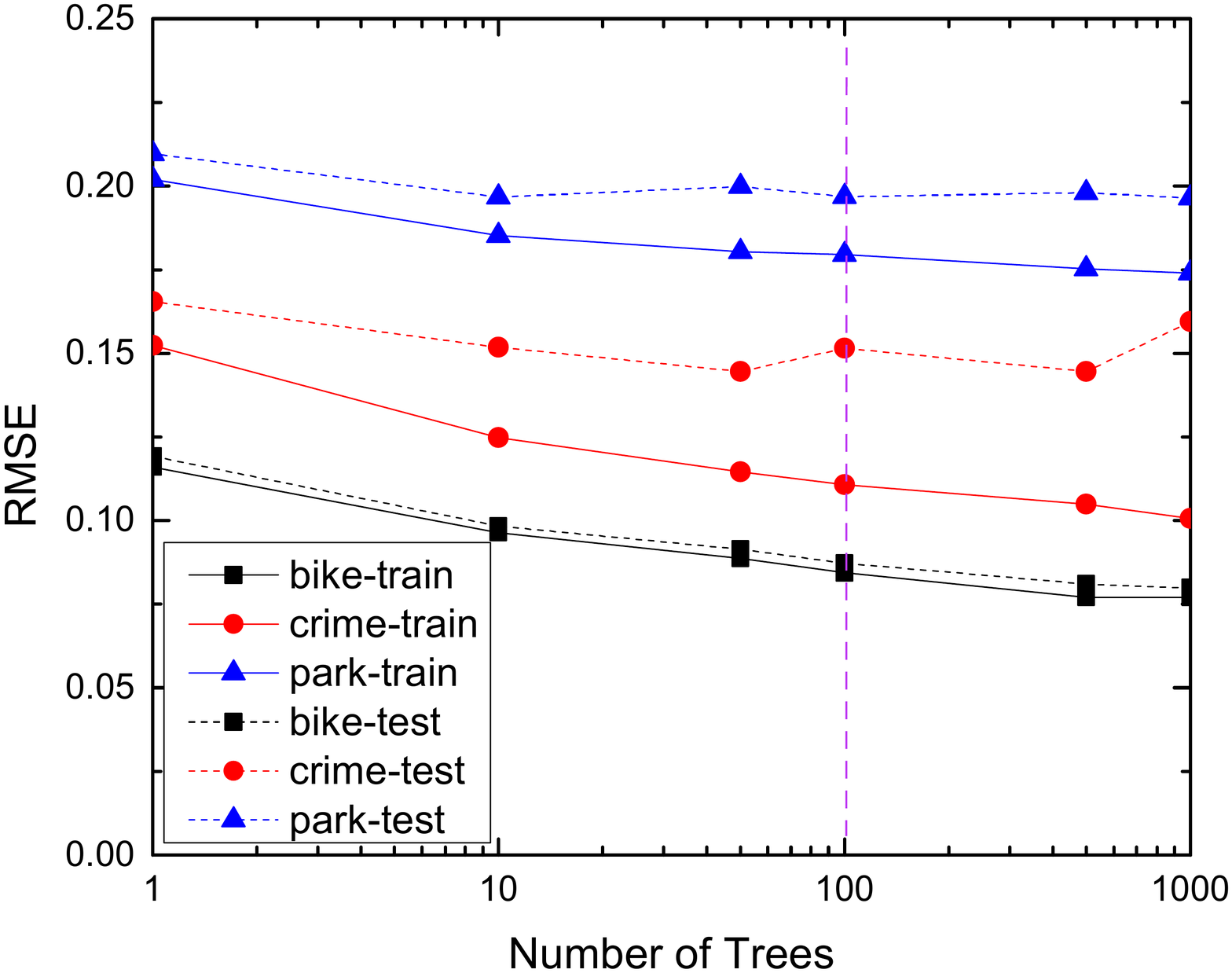}
          }
          \subfigure[Regression: \method-L] {
            \includegraphics[width=0.23\textwidth]{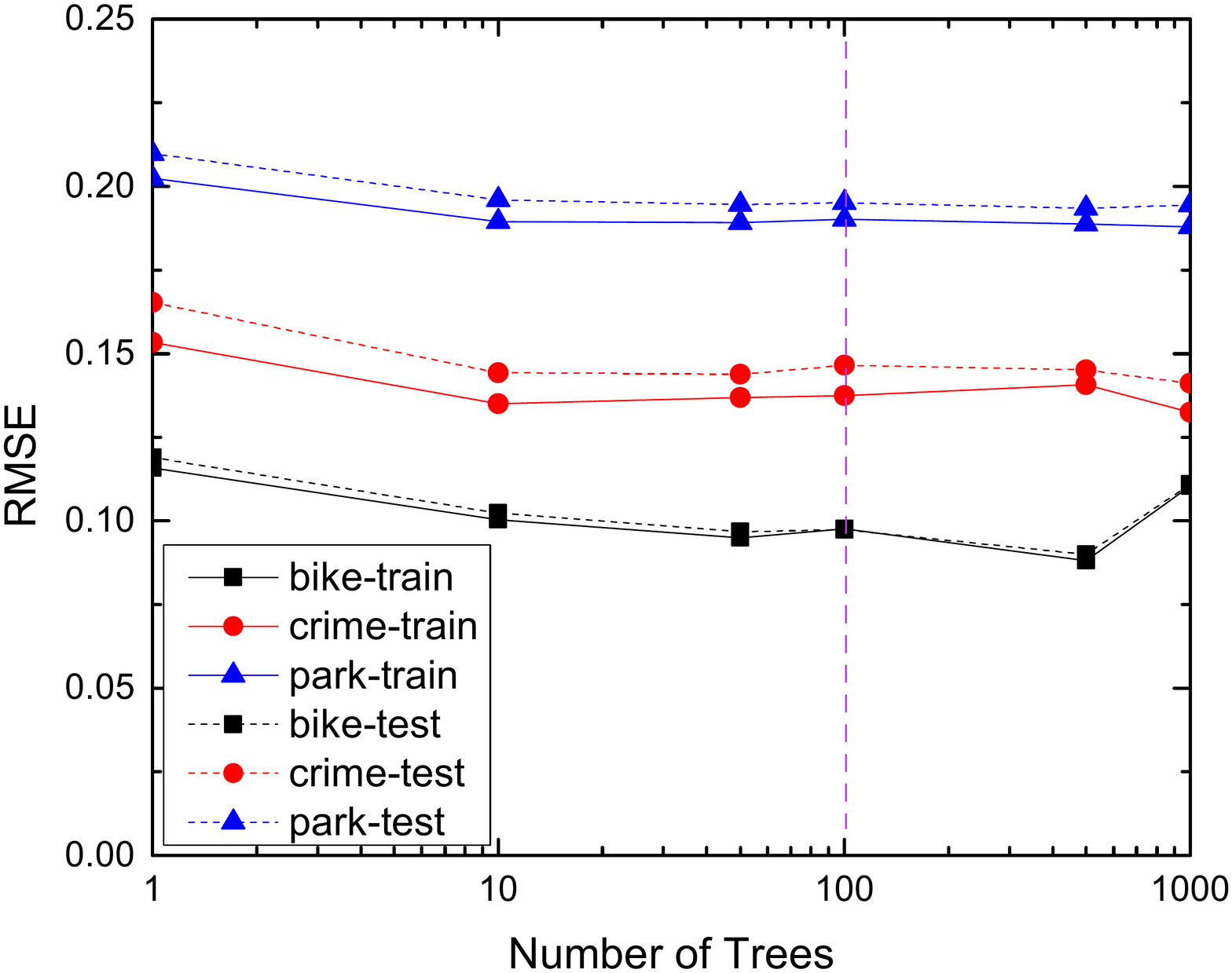}
          }
         \vspace{-0.1in}
          \caption{The impact of the number of trees in classification tasks. Training and testing performances are almost overlapped. We can observe that a small number of trees (e.g., 100) are enough to achieve stable performance.}
          \label{fig:vary_T}
        \vspace{-0.2in}
        \end{figure*}

    \vspace{-0.1in}
    \subsubsection{The Number of Top Discriminative Patterns}

        The most interesting parameter in \method is $k$, the number of discriminative patterns used in the final generalized linear model. It controls the model size of the generalized linear model used for prediction and thus affects its efficiency. Because the default value of $k$ is 20 for classification tasks and 30 for regression tasks and its effectiveness has been proved in previous experiments, we vary $k$ from 1 to 40 to see the trends of both training and testing accuracies on different datasets. Three representative classification datasets (\textit{adult}, \textit{hypo}, and \textit{sick}) and three regression datasets (\textit{bike}, \textit{crime} and \textit{parkinsons}) are used in this experiment.

        As illustrated in Figure~\ref{fig:vary_k}, the performance on test data is always following the trend of performance on training data and the performance is increasing as $k$ grows in both classification and regression tasks (accuracy is increasing on classification datasets while error is decreasing on regression datasets). The discrepancy of training and test performance is more significant in regression tasks (right two in Figure~\ref{fig:vary_k}), which is reasonable due to the higher complexity of the problem, but the trends are quite similar. In addition, we argue that the larger difference could be caused by insufficient size of training data, because the curves always overlap on \textit{bike} dataset that is much bigger than the other two. It is also worth noting that \method-L performs more consistently than \method-F, especially in regression tasks, as a result of $\lambda$ which is automatically learned in \method-L but is manually specified in \method-F. In summary, the similar trends in training and test data justifies that our pattern selection based on training accuracy is reasonable. In real world applications, $k$ could be determined by cross validations.

        Although the performance is becoming better almost all the time, it slows down much when $k$ is greater than the default value. This is true for both classification and regression tasks. An even larger $k$ will hurt the efficiency of both training process and online prediction, and might introduce overfitting issues in prediction (e.g., test accuracy on hypo dataset is 99.58\% when $k=20$ while it becomes 99.28\% when $k=40$ using forward selection). Therefore, we can conclude that a very small $k$ (e.g., $k=20$) is enough for these comprehensive real-world datasets, which further proves that the proposed \method can compress the model into a very tiny size while its accuracy remains comparable.

    \vspace{-0.1in}
    \subsubsection{The Number of Trees in the Model}

        Another important parameter in \method is the number of trees needed to generate the large pool of discriminative patterns. As mentioned before, a single tree is not enough to generate that many patterns, and thus there is strong motivation to try $T=1$ as an extreme case. The default value $100$ works well in previous experiments, and thus we vary $T$ in \{1, 10, 50, 100, 500, 1,000\} to see the trends of both training and testing accuracies. As before, three datasets for classification and regression tasks are presented in the experiments.

        Figure~\ref{fig:vary_T} visualizes the results on classification and regression datasets respectively. When $T=1$, the performance is much lower than others, which means only a single decision tree is not enough for a diverse patterns pool. Too few trees generally cannot guarantee high coverage of effective patterns, especially when data set is large and dimension is high. Increasing number of trees leads to better diversity of candidate patterns. According to the curves, one can easily observe and conclude that the performance remains stable as long as the number of trees is sufficiently large, and a reasonably large $T$ is enough to achieve a satisfying result. Similar to the number of patterns $k$, however, many noisy patterns will be generated if $T$ becomes too large, which fit training data better while fail to characterize testing data and are harmful to generalization of the model (e.g., test RMSE is 0.0977 on hypo dataset when $T=100$ while it becomes 0.1104 when $T=1000$ using LASSO). In addition, the more trees we have, the larger number of pattern candidates will be generated, which increase the time complexity of feature selection. $T$ is by default set to 100 in our experiments, which performs consistently well on different data sets.

\section{Novel Marker Discovery for \mbox{ALS} patient stratification}
\label{sec:app}

    In this section, we apply \method to analyze the prognosis and perform stratification for ALS patients. Unlike other diseases such as many cancers, which can be clearly classified into subtypes with distinct survival rates, no significant signals have been identified to explain the diverse survival times (ranging from less than a year to over 10 years) for ALS patients. Such a wide range makes it difficult to predict disease progression and survival, and suggests rather large underlying disease heterogeneity. There may exist different subgroups of patients, each having its unique disease causes and prognosis.

    \vspace{-0.1in}
    \subsection{ALS Dataset}

        To solve this puzzle, the Pooled Resource Open-Access ALS Clinical Trials (PRO-ACT) platform\footnote{The data in the PRO-ACT Database are contributed by members of PRO-ACT Consortium, founded in 2011 by Prize4Life and the Northeast ALS Consortium with the funding from the ALS Therapy Alliance.} was created by Prize4Life and the Neurological Clinical Research Institute at Massachusetts General Hospital to collect ALS data from existing completed ALS clinical trials. In 2012, a subset of PRO-ACT data was constructed with the aim to crowdsource the challenge of ALS prognosis as a data mining task, which is known as the DREAM-Phil Bowen ALS Prediction Prize4Life Challenge (``\prevchallenge'' for short in this section)~\cite{kuffner2015crowdsourced}.

        \Prevchallenge~aimed at improving the prediction of ALS progression rate, which is essentially a regression task. The participants built models with a training set of 918 patients, and submitted their models to the challenge organizers. The organizers ran the models on a separate leaderboard set of 279 patients and provided feedback on model performance to the participants. Several such submission-and-feedback cycles were run in 3 months, and then the last submissions from the participants were evaluated and ranked by the organizers on another separate validation set of 627 patients.

        This challenge attracted more than 1,000 participants and received 37 unique algorithms during the submission-and-feedback leaderboard phase. Among them, only six algorithms demonstrated improved accuracy over the baseline (developed by the challenge organizers) on the final validation data set.

        The best prognosis model (``\winner'' for short in this section) developed in \prevchallenge, which uses Bayesian trees with 484 predictive features constructed from 26 clinical variables, is a profound success. It has predicted ALS progression from clinical data better than clinicians do, and can potentially reduce the cost of future ALS trials by \$6-million~\cite{kuffner2015crowdsourced}. \Winner~is not perfect though. It is a uniform model for all patients and thus lacks the ability to make personalized diagnosis. Also, it is hard to clinically interpret \winner~due to the high model complexity.

        For fair comparison, \method has been trained and evaluated in such a way that mimics \prevchallenge. Training was performed with the same training set of 918 patients and evaluation was on the same validation set of 627 patients. The leaderboard set of 279 patients was used merely for feature calibration (described later in this paper).

        The data used in \prevchallenge~consist of 2 parts: clinical variables and the actual ALS progression rate (which serves as the golden standard for model comparison). Available clinical variables of a patient can be grouped into 5 kinds: demographic information, vital signs, lab test results, family disease history and the Amyotrophic Lateral Sclerosis Functional Rating Scale~(ALSFRS). A detailed description of the data can be found in the supplement of~\cite{kuffner2015crowdsourced}. Some variables are excluded from our study because their units are not consistent for some patients.

        ALSFRS is a quantitative clinical score ranging from 0 to 40 for evaluating the functional status of an ALS patient. It consists of 10 assessments of motor functioning, each evaluated within the range 0 (worst status, no function) to 4 (normal function). Those 10 evaluated functions\footnote{Some patients are evaluated instead with a modified version ALSFRS-R ranging from 0 to 48, where \mquote{10.respiratory} is replaced with \mquote{R1.Dyspnea}, \mquote{R2.Orthopnea} and \mquote{R3.Respiratory insufficiency}, each ranging from 0 to 4.} are: \mquote{1.speech}, \mquote{2.salivation}, \mquote{3.swallowing}, \mquote{4.handwriting}, \mquote{5.cutting food and handling utensils} (with or without gastrostomy), \mquote{6.dressing and hygiene}, \mquote{7.turning in bed and adjusting bed clothes}, \mquote{8.walking}, \mquote{9.climbing stairs} and \mquote{10.respiratory}.

        The rate of change in ALSFRS\footnote{To assure the consistency of scales across patients, for those patients with ALSFRS-R only but no ALSFRS, the sum of questions $1$-$9$ and $R1$ are used in the calculation of the rate of change.} with respect to time $T$ (\rate) can be used as a quantitative measurement of ALS progression rate. The task is to predict \rate~within 3 to 12 months from disease onset, given the clinical variables within the first 3 months. The RMSE between the predicted \rate~and the actual value is used to evaluate the predictive performance.

    \vspace{-0.1in}
    \subsection{Data Processing}

        The clinical variables about a patient contain 3 data types: static categorical, static continuous and longitudinal continuous variables. Static variables are time-independent, while longitudinal variables are measured multiple times for each patient and are likely to change over time. Any static categorical variable with $k$ categories is replaced with $k$+1 binary features where the additional one indicates whether the variable is missing. A static continuous variable is simply a continuous feature.

        Each longitudinal continuous variable \{$\mathbf{x}$,$\mathbf{t}$\}, where $\mathbf{x} \in \mathbb{R}^n$ is the $n$ measured values and $\mathbf{t} \in \mathbb{R}^n$ is the times of $n$ measurements in ascending order, is converted to 12 continuous features by taking some statistics of $\{\mathbf{x}$,$\mathbf{t}\}$ and a derivative sequence $\mathbf{\Delta} \in \mathbb{R}^{n-1}$ whose $i$th element is defined as $\Delta_i=(x_{i+1}-x_i)/(t_{i+1}-t_i)$. 6 statistics are taken from $\mathbf{x}$: the average value $(\sum_{i=1}^n x_i) / n$, the first-measured value $x_1$, the last-measured value $x_n$, the maximum $\max_i \{ x_i\}$, the minimum $\min_i \{ x_i \}$, and the standard deviation $\sigma(x_i)$. Another 6 statistics are taken similarly from $\mathbf{\Delta}$.

        After performing such variable conversion separately on the training, leaderboard and validation sets, features are calibrated across all 3 sets so that features completely missing in at least 1 of the 3 data sets are discarded. The number of features we finally feed into \method is 498, converted from 78 clinical variables.

    \vspace{-0.1in}
    \subsection{Task Description}
        In the precision medicine setting, we assume there are some implicit groupings underlying the patients, such as the subtypes of a certain disease. Formally, we define the \pc~as follows.
        \begin{thm:def} \textbf{Diagnosis-Stratified Patient Clusters} are $G$ disjoint patient groups, such that patients within the same group are similar and there are different top-$k$ patterns of clinical variables across clusters that suggests distinct diagnoses. We use \textbf{patient cluster} for short in this paper.
        \end{thm:def}

        Considering different patient sets $\mathcal{S}$, we can define the global and local patterns respectively.
        \begin{thm:def} \textbf{Global Patterns} are the top-$K_g$ patterns by using all patients as training instances.
        \end{thm:def}
        The global patterns are expected to not only capture the general properties of the specific task, but also hopefully find the way to detect implicit groups of patients. For example, suppose a disease has 3 different subtypes, we expect some global patterns can handle the general diagnosis while others can help clinicians partition patients into the 3 subtypes.

        \begin{thm:def} \textbf{Local Patterns} are the top-$K_l$ patterns by using only the patients in a single \pc as training instances.
        \end{thm:def}
        Within different \pc s (e.g., different subtypes of a disease), patients may have different root causes, and thus need different diagnoses and treatments. Therefore, we are motivated to discover local patterns.

        In this application, our task is to first discover global patterns for all patients and then figure out the \pc s as well as the local patterns in each \pc. The goal is to demonstrate that our \method can not only accurately predict ALS prognosis, but also systematically identify clinically-relevant features for ALS patient stratification in an interpretable manner, which will further facilitate personalized diagnosis and therapy.

    \vspace{-0.1in}
    \subsection{\method for ALS patient stratification}
        \begin{figure}[t]
          \centering
          \includegraphics[width=3.5in]{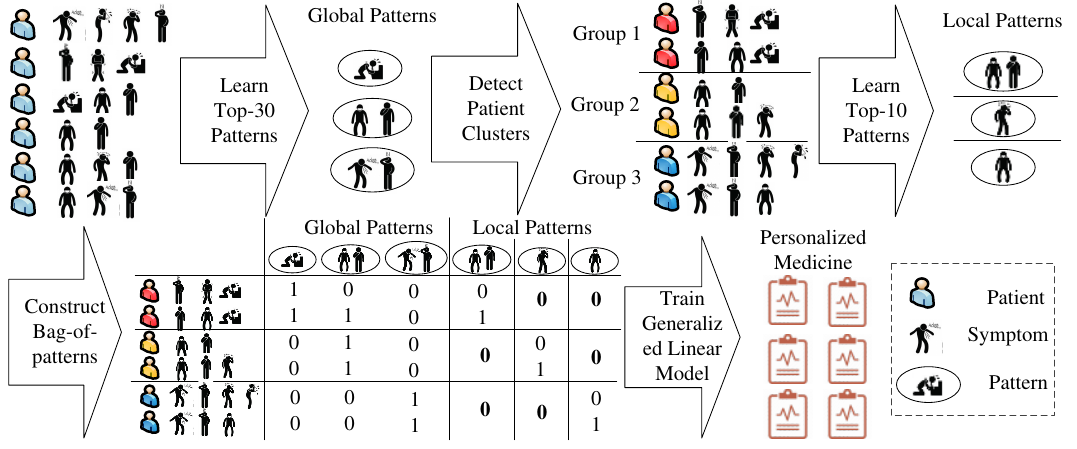}
          \vspace{-0.2in}
          \caption{Overview of the prognosis analysis and stratification for ALS patients. Starting from the training data, \method is first applied to discover global patterns. Second, classical clustering algorithm (e.g., LDA) is utilized to detect diagnosis-Stratified patient clusters in the constructed pattern space. Third, local patterns are explored by \method within a certain patient cluster. In the end, by combining both global and local patterns, a concise and unified generalized linear model is ready for testing data.}\label{fig:overview}
        \vspace{-0.2in}
        \end{figure}

        As shown in Figure~\ref{fig:overview}, the prognosis analysis and stratification for ALS patients work as follows.
        \begin{itemize}
        	\item Discover $K_g$ global patterns based on all patients;
        	\item Partition patients into $G$ different \pc s based on the discovered global patterns;
        	\item Discover $K_l$ local patterns inside each \pc;
        	\item Construct the bag-of-patterns feature representation for each patient based on all global patterns and only the local patterns discovered in his/her \pc;
        	\item Train a generalized linear model based on the constructed features.
        \end{itemize}

        When a new patient comes, it is predicted as follows.
        \begin{itemize}
            \item Assign a \pc based on $K_g$ global patterns;
            \item Evaluate the corresponding $K_l$ local patterns in the assigned \pc;
            \item Construct the bag-of-patterns feature representations based on these $K_g + K_l$ discriminative patterns;
            \item Predict by the generalized linear model.
        \end{itemize}

        We utilize \method to discover global and local patterns. Since it is a regression task, similar to our previous experiments, we set $T = 100, D = 6, \sigma = 10, K_g = 30, K_l = 10$. Therefore, for each patient, we have $K_g + K_l = 40$ patterns. For the patient clustering, by making analogy from bag-of-words to bag-of-patterns, we adopt Latent Dirichlet Allocation (LDA) algorithm~\cite{blei2003LDA} and set $G = 3$. More specifically, observing global patterns of patients, in order to detect \pc s, we design a generative process of the patterns incorporating \pc s as latent variables.
        First, we assume the patterns in a particular \pc follow a multinomial distribution, which is a random variable draws from a prior Dirichlet distribution.
        Inspired from bag-of-words, by making analogies between words in documents and patterns of patients, we represent the observed patterns of a patient as a bag of patterns. Therefore, the generative process can be treated as the process of LDA.

    \vspace{-0.1in}
    \subsection{Results and Discussion}

        \begin{figure}[t]
          \centering
          \includegraphics[width=0.5\textwidth]{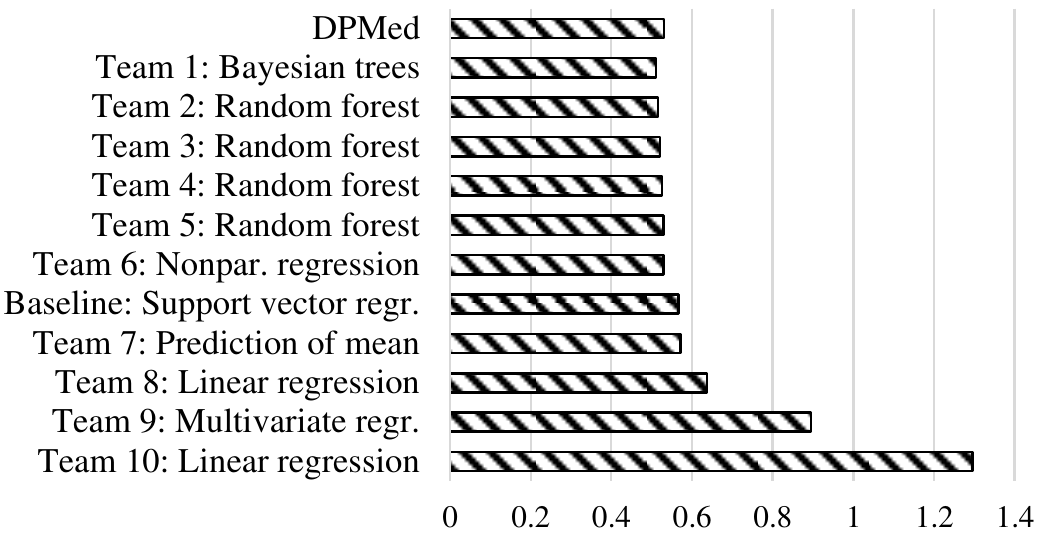}
          \vspace{-0.25in}
          \caption{Testing RMSE in DREAM-Phil Bowen ALS Prediction Prize4Life Challenge. \method has obtained a predictive performance comparable to \winner, which is only $<4\%$ away from the RMSE of \winner, comparable to the other top-ranked algorithms which are also complicated and not interpretable, and better than the baseline RMSE. The linear combination of discriminative patterns trained with \method includes 28 clinical variables, forming a small subset of all 78 variables.}
          \label{fig:performance}
          \vspace{-0.2in}
        \end{figure}

        \method has obtained a predictive performance comparable to \winner~while gives interpretable discriminative patterns, as shown in Figure~\ref{fig:performance}. \method with 3 \pc s achieves a RMSE of $0.5306$ on the validation data set, which is only $<4\%$ away from the RMSE of \winner, $0.5113$, comparable to the other top-ranked algorithms which are also complicated and not interpretable, and better than the baseline RMSE, $0.5664$. The linear combination of discriminative patterns trained with \method includes 28 clinical variables in total, which is a small subset of all 78 available variables.

        Our top 20 most frequent clinical variable list (Figure~\ref{fig:feature_importance}) reveals the importance of the blood urea nitrogen (BUN) and the respiratory rate, which are not among the most important features reported by any of the top 5 teams nor the organizers of \prevchallenge. The other variables in our top 20 list agree well with \prevchallenge~findings. Some examples include the critical role of the onset delta (i.e.the time between the ALS onset and the first time the patient was tested in a trial), mouth-related ALSFRS assessments (including \mquote{1.speech}, \mquote{2.salivation} and \mquote{3.swallowing}) and vital capacity. A high degree of consistency with \prevchallenge~results proves the reliability of \method, while our newly reported important variables highlights the power of feature selection in \method and shed new light on ALS research.

        There are other reasons to take our newly discovered important clinical variables seriously when designing future studies. It has been experimentally shown that the BUN level is elevated ($p < 0.05$) when minocycline, a drug that can delay the progression of ALS, is applied~\cite{zhu2002minocycline,gordon2004placebo}. Therefore the correlation between the BUN level and the ALS progression rate is likely to be true. The respiratory rate reflects respiratory muscle functioning and thus related to \mquote{10.respiratory}, 1 of the 10 assessments in ALSFRS. Since the importance of \mquote{10.respiratory} is reported by several among the top 5 teams in \prevchallenge~\cite{kuffner2015crowdsourced} and also by \method, it should not be surprising that the respiratory rate is also in the list. Interestingly \method is the only algorithm among those that simultaneously selects both the respiratory rate and \mquote{10.respiratory}.

        Another point worth mentioning is the distinct local patterns of each \pc~displayed in Figure~\ref{fig:feature_importance}, indicating different diagnosis patterns across \pc s. For example, the mouth-functioning-related scores are important overall but not locally in Cluster 3, while the blood pressure is important in Patient Clusters 2 \& 3 but plays a less significant role in Cluster 1. Such distinct diagnosis patterns may not only aid personalized medicine but also shed light on the mechanism, underlying heterogeneity and treatment of ALS. To demonstrate the predictive performance of stratification, we also trained a \method model without clustering, and its RMSE, $0.5404$, is worse.

        All these results indicate that our \method not only accurately predicts ALS prognosis, but also systematically identifies clinically-relevant features for ALS patient stratification in an interpretable manner, which will facilitate personalized diagnosis and therapy.

\section{Conclusions}
\label{sec:con}

In this paper, we propose an effective and concise discriminative pattern-based prediction framework (\method) to address the classification and regression problems and provide high interpretability with a small number of discriminative patterns. Specifically, \method first trains a constrained multi-tree model using training data and then extracts the prefix paths from root nodes to non-leaf nodes in all the trees as candidate discriminative patterns. The size of discriminative patterns is compressed by selecting the most effective pattern combinations according to their predictive performance in a generalized linear model. Instead of selecting the patterns independently using heuristics, \method finds the best combination using forward selection or LASSO, which avoids the overlapping effect between similar patterns. Extensive experiments demonstrate that \method is able to model high-order interactions and present a small number of interpretable patterns to help human experts understand the data. \method provides comparable or even better performance than the state-of-the-art model \ddpmine and random forest model in classification and regression. \method has been successfully applied to discover patient clusters and crucial clinical signals for the amyotrophic lateral sclerosis~(ALS) disease.


\vspace{-0.1in}

\bibliographystyle{abbrv}
{
\bibliography{00-paper}

\begin{thebibliography}{10}

\bibitem{blei2003LDA}
D.~M. Blei, A.~Y. Ng, and M.~I. Jordan.
\newblock Latent dirichlet allocation.
\newblock {\em JMLR}, 3:993--1022, 2003.

\bibitem{chen2004using}
C.~Chen, A.~Liaw, and L.~Breiman.
\newblock Using random forest to learn imbalanced data.
\newblock {\em University of California, Berkeley}, 2004.

\bibitem{cheng2007discriminative}
H.~Cheng, X.~Yan, J.~Han, and C.-W. Hsu.
\newblock Discriminative frequent pattern analysis for effective
  classification.
\newblock In {\em Data Engineering, 2007. ICDE 2007. IEEE 23rd International
  Conference on}, pages 716--725. IEEE, 2007.

\bibitem{cheng2008direct}
H.~Cheng, X.~Yan, J.~Han, and P.~S. Yu.
\newblock Direct discriminative pattern mining for effective classification.
\newblock In {\em Data Engineering, 2008. ICDE 2008. IEEE 24th International
  Conference on}, pages 169--178. IEEE, 2008.

\bibitem{cong2005mining}
G.~Cong, K.-L. Tan, A.~K. Tung, and X.~Xu.
\newblock Mining top-k covering rule groups for gene expression data.
\newblock In {\em Proceedings of the 2005 ACM SIGMOD international conference
  on Management of data}, pages 670--681. ACM, 2005.

\bibitem{derksen1992backward}
S.~Derksen and H.~Keselman.
\newblock Backward, forward and stepwise automated subset selection algorithms:
  Frequency of obtaining authentic and noise variables.
\newblock {\em British Journal of Mathematical and Statistical Psychology},
  45(2):265--282, 1992.

\bibitem{deshpande2005frequent}
M.~Deshpande, M.~Kuramochi, N.~Wale, and G.~Karypis.
\newblock Frequent substructure-based approaches for classifying chemical
  compounds.
\newblock {\em TKDE}, 17(8):1036--1050, 2005.

\bibitem{sdm16_dong_pattern}
G.~Dong and V.~Taslimitehrani.
\newblock Pattern aided classification.
\newblock In {\em Proceedings of 2016 SIAM international conference on Data
  Mining}, 2016.

\bibitem{ebina2011drop}
T.~Ebina, H.~Toh, and Y.~Kuroda.
\newblock Drop: an svm domain linker predictor trained with optimal features
  selected by random forest.
\newblock {\em Bioinformatics}, 27(4):487--494, 2011.

\bibitem{fan2008direct}
W.~Fan, K.~Zhang, H.~Cheng, J.~Gao, X.~Yan, J.~Han, P.~Yu, and O.~Verscheure.
\newblock Direct mining of discriminative and essential frequent patterns via
  model-based search tree.
\newblock In {\em Proceedings of the 14th ACM SIGKDD international conference
  on Knowledge discovery and data mining}, pages 230--238. ACM, 2008.

\bibitem{friedman2009glmnet}
J.~Friedman, T.~Hastie, and R.~Tibshirani.
\newblock glmnet: Lasso and elastic-net regularized generalized linear models.
\newblock {\em R package version}, 1, 2009.

\bibitem{friedman2001greedy}
J.~H. Friedman.
\newblock Greedy function approximation: a gradient boosting machine.
\newblock {\em Annals of statistics}, pages 1189--1232, 2001.

\bibitem{ganjisaffar2011bagging}
Y.~Ganjisaffar, R.~Caruana, and C.~V. Lopes.
\newblock Bagging gradient-boosted trees for high precision, low variance
  ranking models.
\newblock In {\em Proceedings of the 34th international ACM SIGIR conference on
  Research and development in Information Retrieval}, pages 85--94. ACM, 2011.

\bibitem{gordon2004placebo}
P.~Gordon, D.~Moore, D.~Gelinas, C.~Qualls, M.~Meister, J.~Werner, M.~Mendoza,
  J.~Mass, G.~Kushner, and R.~Miller.
\newblock Placebo-controlled phase i/ii studies of minocycline in amyotrophic
  lateral sclerosis.
\newblock {\em Neurology}, 62(10):1845--1847, 2004.

\bibitem{hosmer2004applied}
D.~W. Hosmer~Jr and S.~Lemeshow.
\newblock {\em Applied logistic regression}.
\newblock John Wiley \& Sons, 2004.

\bibitem{kobetski2011discriminative}
M.~Kobetski and J.~Sullivan.
\newblock Discriminative tree-based feature mapping.
\newblock {\em Intelligence}, 34(3), 2011.

\bibitem{krizhevsky2012imagenet}
A.~Krizhevsky, I.~Sutskever, and G.~E. Hinton.
\newblock Imagenet classification with deep convolutional neural networks.
\newblock In {\em Advances in neural information processing systems}, pages
  1097--1105, 2012.

\bibitem{kudo2004application}
T.~Kudo, E.~Maeda, and Y.~Matsumoto.
\newblock An application of boosting to graph classification.
\newblock In {\em Advances in neural information processing systems}, pages
  729--736, 2004.

\bibitem{kuffner2015crowdsourced}
R.~K{\"u}ffner, N.~Zach, R.~Norel, J.~Hawe, D.~Schoenfeld, L.~Wang, G.~Li,
  L.~Fang, L.~Mackey, O.~Hardiman, et~al.
\newblock Crowdsourced analysis of clinical trial data to predict amyotrophic
  lateral sclerosis progression.
\newblock {\em Nature biotechnology}, 33(1):51--57, 2015.

\bibitem{lee2006information}
C.~Lee and G.~G. Lee.
\newblock Information gain and divergence-based feature selection for machine
  learning-based text categorization.
\newblock {\em Information processing \& management}, 42(1):155--165, 2006.

\bibitem{leslie2002spectrum}
C.~S. Leslie, E.~Eskin, and W.~S. Noble.
\newblock The spectrum kernel: A string kernel for svm protein classification.
\newblock In {\em Pacific symposium on biocomputing}, volume~7, pages 566--575.
  World Scientific, 2002.

\bibitem{li2001cmar}
W.~Li, J.~Han, and J.~Pei.
\newblock Cmar: Accurate and efficient classification based on multiple
  class-association rules.
\newblock In {\em Data Mining, 2001. ICDM 2001, Proceedings IEEE International
  Conference on}, pages 369--376. IEEE, 2001.

\bibitem{lodhi2002text}
H.~Lodhi, C.~Saunders, J.~Shawe-Taylor, N.~Cristianini, and C.~Watkins.
\newblock Text classification using string kernels.
\newblock {\em JMLR}, 2:419--444, 2002.

\bibitem{lou2012intelligible}
Y.~Lou, R.~Caruana, and J.~Gehrke.
\newblock Intelligible models for classification and regression.
\newblock In {\em Proceedings of the 18th ACM SIGKDD international conference
  on Knowledge discovery and data mining}, 2012.

\bibitem{lou2013accurate}
Y.~Lou, R.~Caruana, J.~Gehrke, and G.~Hooker.
\newblock Accurate intelligible models with pairwise interactions.
\newblock In {\em Proceedings of the 19th ACM SIGKDD international conference
  on Knowledge discovery and data mining}, pages 623--631. ACM, 2013.

\bibitem{ma1998integrating}
B.~L. W. H.~Y. Ma.
\newblock Integrating classification and association rule mining.
\newblock In {\em Proceedings of the fourth international conference on
  knowledge discovery and data mining}, 1998.

\bibitem{moosmann2007fast}
F.~Moosmann, B.~Triggs, and F.~Jurie.
\newblock Fast discriminative visual codebooks using randomized clustering
  forests.
\newblock In {\em Twentieth Annual Conference on Neural Information Processing
  Systems (NIPS'06)}, pages 985--992. MIT Press, 2007.

\bibitem{ren2015global}
S.~Ren, X.~Cao, Y.~Wei, and J.~Sun.
\newblock Global refinement of random forest.
\newblock In {\em Proceedings of the IEEE Conference on Computer Vision and
  Pattern Recognition}, pages 723--730, 2015.

\bibitem{suykens1999least}
J.~A. Suykens and J.~Vandewalle.
\newblock Least squares support vector machine classifiers.
\newblock {\em Neural processing letters}, 9(3):293--300, 1999.

\bibitem{tibshirani1996regression}
R.~Tibshirani.
\newblock Regression shrinkage and selection via the lasso.
\newblock {\em Journal of the Royal Statistical Society. Series B
  (Methodological)}, pages 267--288, 1996.

\bibitem{veloso2006lazy}
A.~Veloso, W.~Meira, and M.~J. Zaki.
\newblock Lazy associative classification.
\newblock In {\em Data Mining, 2006. ICDM'06. Sixth International Conference
  on}, pages 645--654. IEEE, 2006.

\bibitem{vens2011random}
C.~Vens and F.~Costa.
\newblock Random forest based feature induction.
\newblock In {\em Data Mining (ICDM), 2011 IEEE 11th International Conference
  on}, pages 744--753. IEEE, 2011.

\bibitem{wang2005harmony}
J.~Wang and G.~Karypis.
\newblock Harmony: Efficiently mining the best rules for classification.
\newblock In {\em Proceedings of 2005 SIAM international conference on Data
  Mining}, volume~5, pages 205--216. SIAM, 2005.

\bibitem{yin2003cpar}
X.~Yin and J.~Han.
\newblock Cpar: Classification based on predictive association rules.
\newblock In {\em Proceedings of 2003 SIAM international conference on Data
  Mining}, volume~3, pages 369--376. SIAM, 2003.

\bibitem{zhu2002minocycline}
S.~Zhu, I.~G. Stavrovskaya, M.~Drozda, B.~Y. Kim, V.~Ona, M.~Li, S.~Sarang,
  A.~S. Liu, D.~M. Hartley, S.~Gullans, et~al.
\newblock Minocycline inhibits cytochrome c release and delays progression of
  amyotrophic lateral sclerosis in mice.
\newblock {\em Nature}, 417(6884):74--78, 2002.

\end{thebibliography}
}

\begin{IEEEbiography}
[{\includegraphics[width=1in,height=1.25in,clip,keepaspectratio]{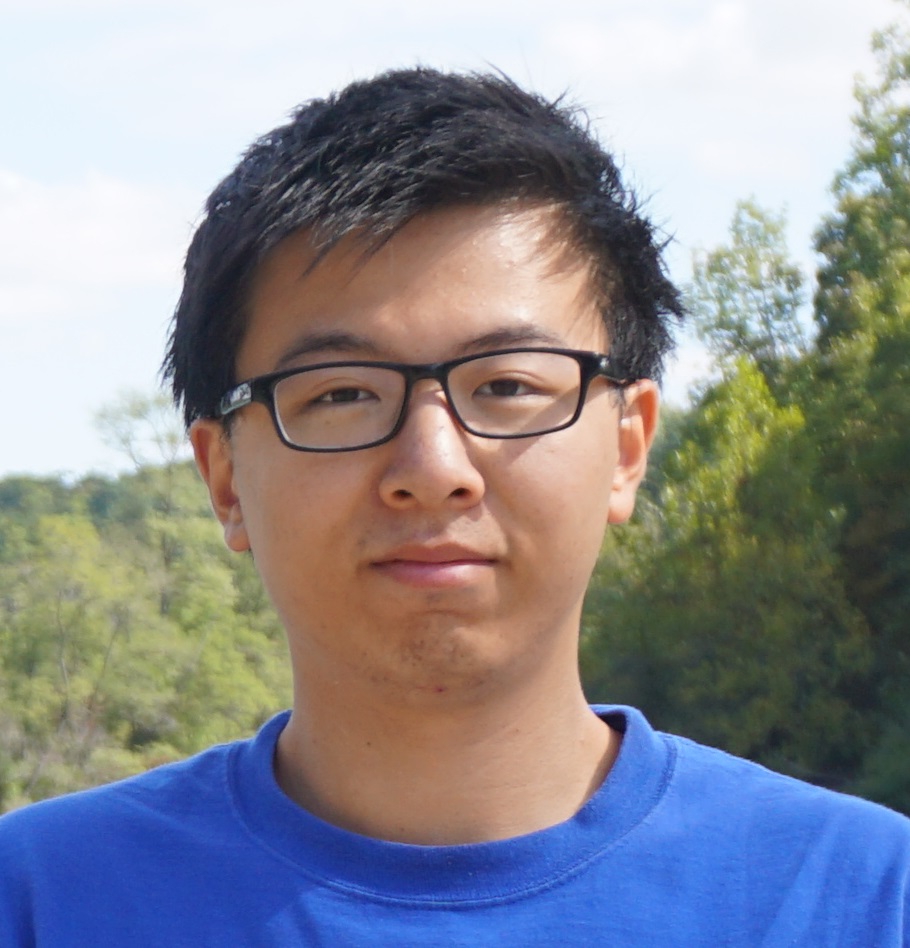}}]
{Jingbo Shang}
received his B.E. degree in 2014 at the Department of Computer Science and Engineering in Shanghai Jiao Tong University, China.
He is now a Ph.D. candidate at University of Illinois at Urbana-Champaign, supervised by Prof. Jiawei Han.
His main research interests include large-scale data mining and data mining, as well as constructing and mining heterogeneous information network from massive text corpora.
\end{IEEEbiography}

\vspace{-0.5in}
\begin{IEEEbiography}
[{\includegraphics[width=1in,height=1.25in,clip,keepaspectratio]{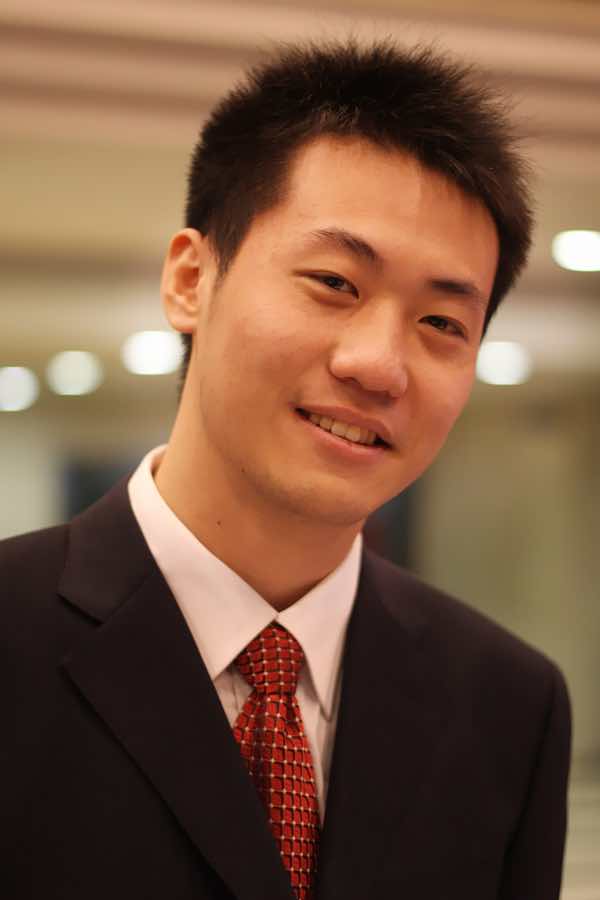}}]
{Meng Jiang}
received his B.E. degree and Ph.D. degree in 2010 and 2015
at the Department of Computer Science and Technology in Tsinghua University.
He is now a Postdoctoral Research Associate at University of Illinois at Urbana-Champaign.
He visited Carnegie Mellon University from 2012 to 2013.
He has over 15 published papers on Data-Driven Behavioral Analytics for
recommendation, prediction and suspicious behavior detection
in top conferences and journals of the relevant field such as IEEE TKDE, ACM SIGKDD, AAAI, ACM CIKM and IEEE ICDM.
He got the best paper finalist in ACM SIGKDD 2014.
\end{IEEEbiography}

\vspace{-0.5in}
\begin{IEEEbiography}
[{\includegraphics[width=1in,height=1.25in,clip,keepaspectratio]{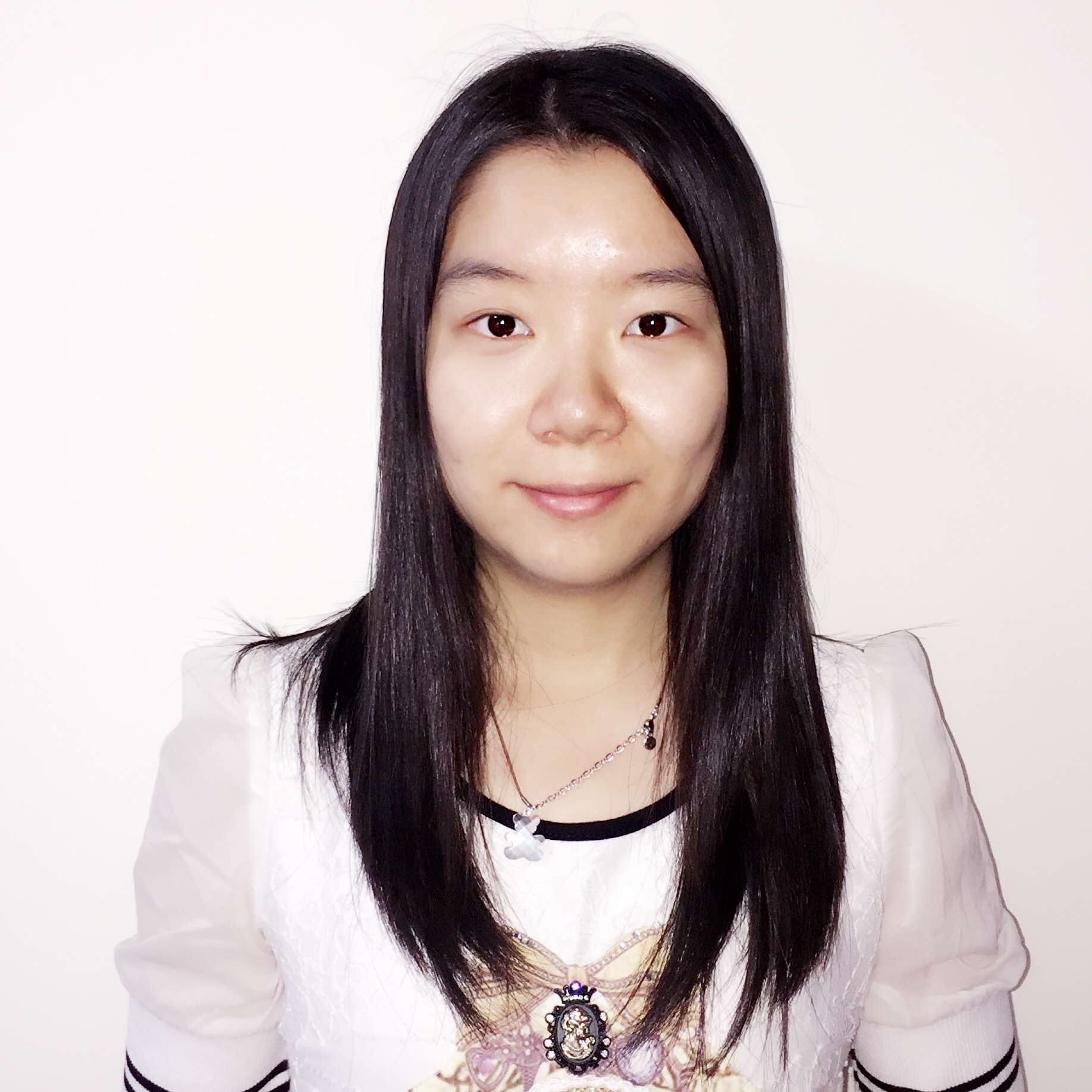}}]
{Wenzhu Tong}
received her B.E. degree in 2014 at the Department of Computer Science and Technology in Wuhan University, China and M.S. degree in 2016 at the Computer Science Department in University of Illinois, Urbana-Champaign.
\end{IEEEbiography}

\vspace{-0.5in}
\begin{IEEEbiography}
[{\includegraphics[width=1in,height=1.25in,clip,keepaspectratio]{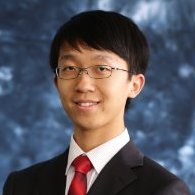}}]
{Jinfeng Xiao}
received his B.S. degree in Physics and Mathematics in 2014 at the Hong Kong University of Science and Technology. He is now a graduate student in Computer Science at the University of Illinois at Urbana-Champaign. His current research lies at the interface of machine learning, data mining and biomedical science.
\end{IEEEbiography}

\vspace{-0.5in}
\begin{IEEEbiography}
[{\includegraphics[width=1in,height=1.25in,clip,keepaspectratio]{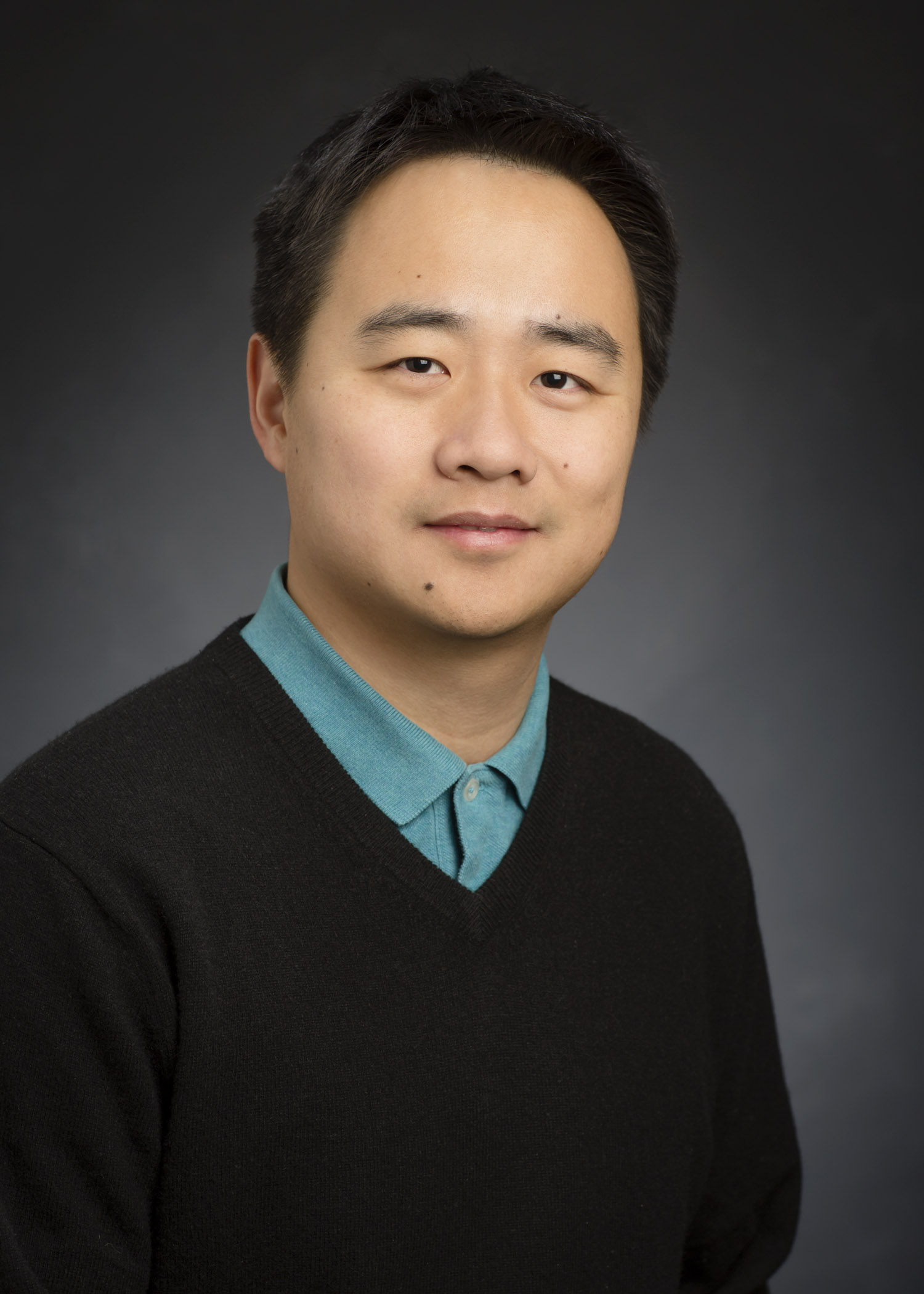}}]
{Jian Peng}
is an assistant professor in the Department of Computer Science at the University of Illinois at Urbana-Champaign. Before joining CS@Illinois in 2015, Jian was a postdoc in the Berger Lab at MIT, and a visiting scientist in the Lindquist Lab at the Whitehead Institute for Biomedical Research. He was a student in the Xu Lab from 2007 and obtained his Ph.D. in Computer Science from Toyota Technological Institute at Chicago in 2013. He received his B.S. in Computer Science from Wuhan University. Jian worked on HIV protein analysis with Drs. David Heckerman and Jonathan Carlson in the eScience group at Microsoft Research in 2010.
\end{IEEEbiography}

\vspace{-0.5in}
\begin{IEEEbiography}
[{\includegraphics[width=1in,height=1.25in,clip,keepaspectratio]{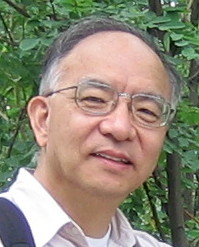}}]
{Jiawei Han}
is Abel Bliss Professor in the Department of Computer Science at the University of Illinois. He has been researching into
data mining, information network analysis, and database systems, with over 600 publications. He served as the founding Editor-in-Chief
of ACM Transactions on Knowledge Discovery from Data (TKDD). Jiawei has received ACM SIGKDD Innovation Award (2004), IEEE Computer Society Technical Achievement Award (2005), IEEE Computer Society W. Wallace McDowell Award (2009), and Daniel C. Drucker Eminent Faculty Award at UIUC (2011). He is a Fellow of ACM and a Fellow of IEEE. He is currently the Director of Information Network Academic Research Center (INARC) supported by the Network Science-Collaborative Technology Alliance (NS-CTA) program of U.S. Army Research Lab. His co-authored textbook ``Data Mining: Concepts and Techniques'' (Morgan Kaufmann) has been adopted worldwide.
\end{IEEEbiography}
\end{document}